\documentclass[12pt]{article}
\usepackage[draft]{changes}
\usepackage{inputenc}
\usepackage{amssymb}
\usepackage{graphicx}
\usepackage{caption}
\usepackage{subcaption}
\usepackage{amsmath}
\usepackage{url}
\usepackage{float}
\usepackage{epsfig}
\usepackage{setspace}

\newcommand{\be}{\begin{equation}}
\newcommand{\ee}{\end{equation}}

\definechangesauthor[color=red]{zl}

\begin{document}
\date{}
\title{Learning in the Machine:\\ Random Backpropagation and the Deep Learning Channel}
\author{ {Pierre Baldi$^{1,}${\footnote{Corresponding author.$^1$ Department of Computer Science, University of California, Irvine. $^2$ Department of Mathematics, University of California, Irvine.}}, Peter Sadowski$^1$, and Zhiqin Lu$^2$}}
\maketitle

\begin{abstract}
{\bf Abstract:} Random backpropagation (RBP) is a variant of the backpropagation algorithm for training neural networks, where the transpose of the forward matrices are replaced by fixed random matrices in the calculation of the weight updates. It is remarkable both because of its effectiveness, in spite of using random matrices to communicate error information, and because it completely removes the taxing requirement of maintaining symmetric weights in a physical neural system. To better understand random backpropagation, we first connect it to the notions of local learning and  learning channels. Through this connection, we derive several alternatives to RBP, including skipped RBP (SRPB), adaptive RBP (ARBP), sparse RBP, and their combinations (e.g. ASRBP) and analyze their computational complexity. We then study their behavior through simulations using the MNIST and CIFAR-10 bechnmark datasets. These simulations show that most of these variants work robustly, almost as well as backpropagation, and that multiplication by the derivatives of the activation functions is important. As a follow-up, we study also the low-end of the number of bits required to communicate error information over the learning channel. We then provide partial intuitive explanations for some of the remarkable properties of RBP and its variations. Finally, we prove several mathematical results, including the convergence to fixed points of linear chains of arbitrary length, the convergence to fixed points of linear autoencoders with decorrelated data, the long-term existence of solutions for linear systems with a single hidden layer and convergence in special cases, and the convergence to fixed points of non-linear chains, when the derivative of the activation functions is included.
\end{abstract}

\section{Introduction}

Over the years, the question of biological plausibility of the backpropagation algorithm, implementing stochastic gradient descent in neural networks, has been raised several times. The question has gained further relevance due to the numerous successes achieved by backpropagation in a variety of problems
ranging from
computer vision 
\cite{krizhevsky2012imagenet,szegedy2015going,srivastava2015training,he2015deep} to speech recognition \cite{graves2013speech} in engineering, and from high energy physics \cite{baldi2014searching,baldidarkmatter15}
to biology \cite{deepcontact2012,zhou2015predicting,baldiagostinelli2016} in the natural sciences, 
as well to recent 
results on the optimality of backpropagation 
\cite{baldi2016local}.
There are however, several well known issues facing biological neural networks in relation to backpropagation, these include:
(1) the continuous real-valued nature of the gradient information and its ability to change sign, violating Dale's Law;
(2) the need for some kind of teacher's signal to provide targets;
(3) the need for implementing all the linear operations involved in backpropagation;
(4) the need for multiplying the backpropagated signal by the derivatives of the forward activations each time a layer is traversed;
(5) the need for precise alternation between forward and backward passes;
and (6) the complex geometry of biological neurons and the problem of transmitting error signals with precision down to individual synapses.
However, perhaps the most formidable obstacle is that the standard backpropagation algorithm requires propagating error signals backwards using synaptic weights that are {\it identical} to the corresponding forward weights. Furthermore, a related problem that has not been sufficiently recognized, is that this weight symmetry must be maintained at all times during learning, and not just during early neural development. It is hard to imagine mechanisms by which biological neurons could both create and maintain such perfect symmetry.
However, recent simulations \cite{lillicrap2014random}
surprisingly indicate that such symmetry may not be required after all, and that in fact backpropagation works more or less as well when {\it random} weights are used to backpropagate the errors. Our general goal here is to investigate backpropagation with random weights and better understand why it works. 

The foundation for better understanding random backpropagation (RBP) is provided by the concepts of local learning and deep learning channels introduced in \cite{baldi2016local}. Thus we begin by introducing the notations and connecting RBP to these concepts. In turn, this leads to the derivation of several alternatives to RBP, which we study through simulations on well known benchmark datasets before proceeding with more formal analyses.

\section{Setting, Notations, and the Learning Channel}

Throughout this paper, we consider layered feedforward neural networks and supervised learning tasks. We will denote such an architecture by

\be
{\cal A}[N_0,\ldots, N_h, \ldots ,N_L]
\label{eq:arc}
\ee
 where $N_0$ is the size of the input layer, $N_h$ is the size of hidden layer $h$, and $N_L$ is the size of the output layer. We assume that the layers are fully connected and let $w^h_{ij}$ denote the weight connecting neuron $j$ in layer $h-1$ to neuron $i$ in layer $h$. The output
$ O_i^h $ of neuron $i$ in layer $h$ is computed by:

\be
O_i^h=f_i^h(S_i^h) \quad {\rm where} \quad S_i^h=\sum_j w_{ij}^h O^{h-1}_j
\label{eq:neuron}
\ee
The transfer functions $f_i^h$ are usually the same for most neurons, with typical exceptions for the output layer, and usually are monotonic increasing functions.
The most typical functions used in artificial neural networks are the: identity, logistic, hyperbolic tangent, rectified linear, and softmax.

We assume that there is a training set of $M$ examples consisting of input and output-target pairs $(I(t), T(t))$, with $t=1,\ldots, M$. $I_i(t)$ refers to the $i$-th component of the $t$-th input training example, and similarly for the target $T_i(t)$. In addition, there is an error function $\cal E$ to be minimized by the learning process. In general we will asssume standard error functions such as the squared error in the case of regression and identity transfer functions in the output layer, or relative entropy in the case of classification with logistic (single class) or softmax (multi-class) units in the output layer, although this is not an essential point.

While we focus on supervised learning, it is worth noting that several ``unsupervised'' learning algorithms for neural networks (e.g. autoencoders, neural autoregressive distribution estimators, generative adversarial networks) come with output targets and thus fall into the framework used here.  

\subsection{Standard Backpropagation (BP)} 

Standard backpropagation implements gradient descent on $\cal E$, and can be applied in a stochastic fashion on-line (or in mini batches) or in batch form, by summing or averaging over all training examples.
For a single example, omitting the $t$ index for simplicity, the standard backpropagation learning rule is easily obtained by applying the chain rule and given by:

\be
\Delta w_{ij}^h=-\eta \frac{ \partial {\cal E}}{\partial w_{ij}^h}=\eta B_i^hO_j^{h-1}
\label{eq:bp}
\ee
where $\eta$ is the learning rate, $O_j^{h-1}$ is the presynaptic activity, and 
$B_i^h$ is the backpropagated error. Using the chain rule, it is easy to see that the backpropagated error satisfies the recurrence relation:

\be
B_i^h=\frac { \partial {\cal E}}{\partial S^h_i}=(f_i^h)' \sum_k B^{h+1}_k w^{h+1}_{ki}
\label{eq:bp}
\ee
with the boundary condition:

\be
B_i^L=\frac{\partial {\cal E}_i}{\partial S^L_i}=T_i-O^L_i
\label{eq:bp1}
\ee
Thus in short the errors are propagated backwards in an essentially linear fashion using the transpose of the forward matrices, hence the symmetry of the weights, with a multiplication by the derivative of the corresponding forward activations every time a layer is traversed.

\subsection{Standard Random Backpropagation (RBP)}

Standard random backpropagation operates exactly like backpropagation except that the weights used in the backward pass are completely random and fixed. Thus the learning rule becomes:

\be
\Delta w_{ij}^h=\eta R_i^h O_j^{h-1}
\label{eq:bp}
\ee
where the randomly back-propagated error satisfies the recurrence relation:

\be
R_i^h= (f_i^h)' \sum_k R^{h+1}_k c^{h+1}_{ki}
\label{eq:bp}
\ee
and the weights $c^{h+1}_{ki}$ are random and fixed.
The boundary condition at the top remains the same:

\be
R_i^L=\frac{\partial {\cal E}_i}{\partial S^L_i}=T_i-O^L_i
\label{eq:bp1}
\ee
Thus in RBP the weights in the top layer of the architecture are updated by gradient descent, identically to the BP case.

\subsection{The Critical Equations}

Within the supervised learning framework considered here, the goal is to find an optimal set of weights $w_{ij}^h$. The equations that the weights must satisfy at any critical point are simply:

\be
\frac{\partial {\cal E}}{\partial w_{ij}^h}=\sum_t B_i^h(t)O_j^{h-1}(t)=0
\label{eq:critical}
\ee
Thus in general the optimal weights must depend on both the input and the targets, as well as the other weights in the network. And learning can be viewed as a lossy storage procedure for transferring the information contained in the training set into the weights of the architecture.

The critical Equation \ref {eq:critical} shows that all the necessary forward information about the inputs and the lower weights leading up to layer $h-1$ is subsumed by the term $O_j^{h-1}(t)$. Thus in this framework a separate channel for communicating information about the inputs to the deep weights is not necessary. Thus here we focus on the feedback information about the targets, contained in the term $B^h_i(t)$ which, in a physical neural system, must be transmitted through a dedicated channel. 

Note that $B^h_i(t)$ depends on the output $O^L(t)$, the target $T(t)$, as well as all the weights in the layers above $h$ in the fully connected case (otherwise just those weight which are on a path from unit $i$ in layer $h$ to the output units), and in two ways: through $O^L(t)$ and through the backpropagation process. In addition,  $B^h_i(t)$  depends also on all the upper derivatives, i.e. the derivatives of the activations functions for all the neurons above unit $i$ in layer $h$ in the fully connected case (otherwise just those derivatives which are on a path from unit $i$ in layer $h$ to the output units).
Thus in general, in a solution of the critical equations, the weights $w_{ij}^h$ {\it must} depend on $O_j^{h-1}$, the outputs, the targets, the upper weights, and the upper derivatives. 
Backpropagation shows that it is sufficient for the weights to depend on $O_j^{h-1}$, $T-O$, the upper weights, and the upper derivatives.

\subsection{Local Learning}

Ultimately, for optimal learning, all the information required to reach a critical point of $\cal E$ must appear in the learning rule of the deep weights.
In a physical neural system, learning rules must also be local \cite{baldi2016local}, in the sense that they can only involve variables that are available locally in both space and time, although for simplicity here we will focus only on locality in space. 
Thus typically, in the present formalism, a local learning rule for a deep layer must be of the form

\be
\Delta w_{ij}^h=F(O_i^h,O_j^{h-1},w_{ij}^h)
\label{eq:local1}
\ee
and

\be
\Delta w_{ij}^L=F(T_i,O_i^L,O_j^{L-1},w_{ij}^L)
\label{eq:local2}
\ee
assuming that the targets are local variables for the top layer. Among other things, this allows one to organize and stratify learning rules, for instance by considering polynomial learning rules of degree one, two, and so forth. 

Deep local learning is the term we use to describe the use of local learning in all the adaptive layers of a 
feedforward architecture. Note that Hebbian learning \cite{hebb1949organization} is a form of local learning, and deep local learning has been proposed for instance by Fukushima \cite{fukushima1980neocognitron} to train the neocognitron architecture, essentially a feed forward convolutional neural network inspired by 
the earlier neurophysiological work of Hubel and Wiesel \cite{hubel1962receptive}. However, in deep local learning, information about the targets is not propagated to the deep layers and therefore in general deep local learning cannot
find solutions of the critical equations, and thus cannot succeed at learning complex functions \cite{baldi2016local}. 

\subsection{The Deep Learning Channel}

From the critical equations, any optimal neural network learning algorithm must be capable of communicating some information about the outputs, the targets, and the upper weights to the deep weights and, in a physical neural system, a communication channel \cite{Shannon:48,Shannon:48-3} must exist to communicate this information. This is the deep learning channel, or {\it learning channel} in short \cite{baldi2016local}, which can be studied using tools from information and complexity theory.
In physical systems the learning channel must correspond to a physical channel and this leads to important considerations regarding its nature, for instance whether it uses the forward connections in the reverse direction or a different set of connections. Here, we focus primarily on how information is coded and sent over this channel. 

In general, the information about the outputs and the targets communicated through this channel to $w_{ij}^h$ is denoted by $I_{ij}^h(T,O^L)$.
Although backpropagation propagates this information from the top layer to the deep layers in a staged way, this is not necessary and $I_{ij}^h(T,O^L)$ could be sent directly to the deep layer $h$ somehow skipping all the layers above.
This observation leads immediately to the skipped variant of RBP described in the next section. 
It is also important to note that in principle this information should have the form
$I_{ij}^h(T,O^L, w_{rs}^l \; {\rm for} \; l>h, f'(S_r^l)\; {\rm for} \; l \geq h) $. However standard backpropagation shows that it is possible to send the same information to all the synapses impinging onto the same neuron, and thus it is possible to learn with a simpler type of information of the form
$I_{i}^h(T,O^L, w_{rs}^l \; {\rm for} \; l>h,  f'(S_r^l)\; {\rm for} \; l \geq h)  $ targeting the postsynaptic neuron $i$.
This class of algorithms or channels is what we call deep targets algorithms, as they are equivalent to providing a target for each deep neuron.
Furthermore, backpropagation shows that all the necessary information about the outputs and the targets is contained in the term $T-O^L$ so that we only need 
$I_{i}^h(T-O^L, w_{rs}^l \; {\rm for} \; l \geq h, f'(S_r^l)\; {\rm for} \; l>h)$.
Standard backpropagation uses information about the upper weights in two ways: (1)through the output $O^L$ which appears in the error terms $T-O^L$; and through the backpropagation process itself.
{\it Random backpropagation crucially shows that the information about the upper weights contained in the backpropagation process is not necessary}. Thus ultimately we can focus exclusively on information which has the simple form:
$I_{i}^h(T-O^L, r_{rs}^l \; {\rm for} \; l \geq h, f'(S_r^l)\; {\rm for} \; l \geq h)$, where $r$ denotes a set of fixed random weights. 

Thus, using the learning channel, we are interested in  local learning rules of the form:

\be
\Delta w_{ij}^h=F(O_i^h,O_j^{h-1},w_{ij}^h,I_{i}^h(T-O^L,r_{rs}^l \; {\rm for} \; l \geq h,
f'(S_r^l)\; {\rm for} \; l \geq h))
\label{eq:local3}
\ee
In fact, here we shall focus exclusively on learning rules with the multiplicative form:

\be
\Delta w_{ij}^h =\eta I_{i}^h(T-O^L,r_{rs}^l \; {\rm for} \; l \geq h,f'(S_r^l)\; {\rm for} \; l \geq h)   O_j^{h-1}
\label{eq:local4}
\ee
corresponding to a product of the presynaptic activity with 
some kind of backpropagated error information, with standard BP and RBP as a special cases. Obvious important questions, for which we will seek full or partial answers, include:
(1) what kinds of forms can $
I_{i}^h(T-O^L,r_{rs}^l \; {\rm for} \; l \geq h,f'(S_r^l)\; {\rm for} \; l \geq h) $ take (as we shall see there are multiple possibilities)? (2) what are the corresponding tradeoffs among these forms, for instance in terms of computational complexity or information transmission?
and (3) are the upper derivatives necessary and why?

\section{Random Backpropagation Algorithms and Their Computational Complexity}

We are going to focus on algorithms where the information required for the deep weight updates 
$I_{i}^h(T-O^L,f'(S_r^l) \;{\rm for} \; l \geq h)$ is produced essentially through a {\it linear} process whereby the vector $T(t) -O(t) $, computed in the output layer, is processed through linear operations, i.e. additions and multiplications by constants (which can include multiplication by the upper derivatives). Standard backpropagation is such an algorithm, but there are many other possible ones. We are interested in the case where the matrices are random. However, even within this restricted setting, there are several possibilities, depending for instance on: (1) whether the information is progressively propagated through the layers (as in the case of BP), or broadcasted directly to the deep layers; (2) whether multiplication by the derivatives of the forward activations is included or not; and (3) the properties of the matrices in the learning channel (e.g. sparse vs dense). This leads to several new algorithms. Here we will use the following notations:
\begin{itemize}
\item BP= (standard)  backpropagation.
\item RBP= random backpropagation, where the transpose of the feedforward matrices are replaced by random matrices.
\item SRBP = skipped random backpropagation, where the
backpropagated signal arriving onto layer $h$ is given by
$C^h(T-O)$ with a random matrix $C^h$ directly connecting the output layer $L$ to layer $h$, and this for each layer $h$.
\item ARBP = adaptive random backpropagation, where the matrices in the learning channel are initialized randomly, and then progressively adapted during learning using the product of the corresponding forward and backward signals, so that $\Delta c_{rs}^l = \eta R_s^{l+1} O_r^{l} $, where $R$ denotes the randomly backpropagated error. In this case, 
the forward channel becomes the learning channel for the backward weights.
\item ASRBP = adaptive skipped random backpropagation, which combines adaptation with skipped random backpropagation.
\item The default for each algorithm involves the multiplication at each layer by the derivative of the forward activation functions. The variants where this multiplication is omitted will be denoted by: ``(no f')''.
\item The default for each algorithm involves dense random matrices, generated for instance by sampling from a normalized  Gaussian for each weight. But one can consider
also the case of random $\pm 1$ (or (0,1)) binary matrices, or other distributions, including sparse versions of the above.
\item As we shall see, using random weights that have the same sign as the forward weights is not essential, but can lead to improvements in speed and stability. Thus we will use the word ``{\it congruent weights}'' to describe this case. Note that with fixed random matrices in the learning channel initialized congruently, congruence can be lost during learning when the sign of a forward weight changes.
\end{itemize}

SRBP is introduced both for information theoretic reasons-- what happens if the error information is communicated directly?--and because it may facilitate the mathematical analyses since it avoids the backpropagation process. However, in one of the next sections, we will also show empirically that SRBP is a viable learning algorithm,
which in practice can work even better than RBP. 
{\it Importantly, these simulation results suggest that when learning the synaptic weight $w_{ij}^h$ the information  about all the upper derivatives ($f'(S_r^l)\; {\rm for} \; l \geq h)$) is not needed. However the immediate ($l=h$) derivative $f'(S_i^h)$ is needed.}

Note this suggests yet another possible algorithm, skipped backropagation (SBP). In this case, for each training example and at each epoch, the matrix used in the feedback channel is the product of the corresponding transposed forward matrices, ignoring multiplication by the derivative of the forward transfer functions in all the layers above the layer under consideration. Multiplication by the derivative of the forward transfer functions is applied to the layer under consideration. Another possibility is to have a combination of RBP and SRBP in the learning channel, implemented by a combination of long-ranged connections carrying SRBP signals with short-range connections carrying a backpropagation procedure, when no long-range signals are available. This may be relevant for biology since combinations of long-ranged and short-ranged feedback connections is common in biological neural systems. 

In general, in the case of linear networks, $f'=1$ and therefore including or excluding derivative terms makes no difference. Furthermore, for any linear architecture
${\cal A}[N,\ldots, N, \ldots ,N]$ where all the layers have the same size, then RBP is equivalent to SRBP.
However, if the layers do not have the same size, then the layer sizes introduce rank constraints on the information that is backpropagated through RBP that may differ from the information propagated through SRBP. In both the linear and non-linear cases, for any network of depth 3 ($L=3$), RBP is equivalent to SRBP, since there is only one random matrix.

Additional variations can be obtained
by using dropout, or multiple sets of random matrices, in the learning channel, for instance for averaging purposes.
Another variation in the skipped case is cascading, i.e. allowing backward matrices in the learning channel between    all pairs of layers. Note that the notion of cascading increases the number of weights and computations, yet it is still interesting from an exploratory and robustness point of view.

\subsection{Computational Complexity Considerations}

The number of computations required to send error information over the learning channel is a fundamental quantity which, however, depends on the computational model used and the cost associated with various operations. Obviously, everything else being equal, the computational cost of BP and RBP are basically the same since they differ only by the value of the weights being used. However more subtle differences can appear with some of the other algorithms, such as SRBP.

To illustrate this, consider an architecture 
${\cal A}[N_0,\ldots, N_h, \ldots ,N_L]$, fully connected, 
and let $W$ be the total number of weights.
In general, the primary cost of BP is the multiplication of each synaptic weight by the corresponding signal in the backward pass. Thus it is easy to see that the bulk of the operations required for BP to compute the backpropagated signals scale like $O(W)$ (in fact $\Theta (W)$) with:

\be
W=N_0 \times N_1+N_1\times N_2 \ldots +N_{L-1} \times N_L=\sum_{k=0}^{L-1}N_kN_{k+1}
\label{eq:}
\ee
Note that whether biases are added separately or, equivalently, implemented by adding a unit clamped to one to each layer, does not change the scaling. Likewise, adding the costs associated with the sums computed by each neuron and the multiplications by the derivatives of the activation functions does not change the scaling, as long as these operations have costs that are within a constant multiplicative factor of the cost for multiplications of signals by synaptic weights. 

As already mentioned, the scaling for RBP is obviously the same, just using different matrices. However the corresponding term for SRBP is given by

\be
W'=N_L \times N_1+ N_L \times N_2 \ldots N_L \times  N_{L-1} = N_L 
\sum_{k=1}^{k=L-1} N_k
\label{eq:}
\ee
In this sense, the computational complexity of BP and SRBP is identical if all the layers have the same size, but it can be significantly different otherwise, especially taking into consideration the tapering off associated with most architectures used in practice. In a classification problem, for instance, $N_L=1$ and all the random matrices in SRBP have rank 1, and $W'$ scales like the total number of neurons, rather than the total number of forward connections.
Thus, provided it leads to effective learning, SRBP could lead to computational savings in a digital computer. However, in a physical neural system, in spite of these savings, the scaling complexity of BP and SRBP could end up being the same. This is because in a physical neural system, once the backpropagated signal has reached neuron $i$ in layer $h$ it still has to be communicated to the synapse.
A physical model would have to specify the cost of such communication. Assuming one unit cost, both BP and SRBP would require $\Theta (W)$ operations across the entire architecture. Finally, a full analysis in a physical system would have to take into account also costs associated with wiring, and possibly differential costs between long and short wires as, for instance, SRBP requires longer wires
than standard BP or RBP. 

\section{Algorithm Simulations}

In this section, we simulate the various algorithms using standard benchmark datasets. The primary focus is not on 
achieving state-of-the-art results, but rather on better understanding these new algorithms and where they break down. The results are summarized in Table \ref{tab:summary} at the end.

\subsection{MNIST}

\begin{figure}[H]
\begin{subfigure}[b]{\textwidth}
\includegraphics[width=\textwidth]{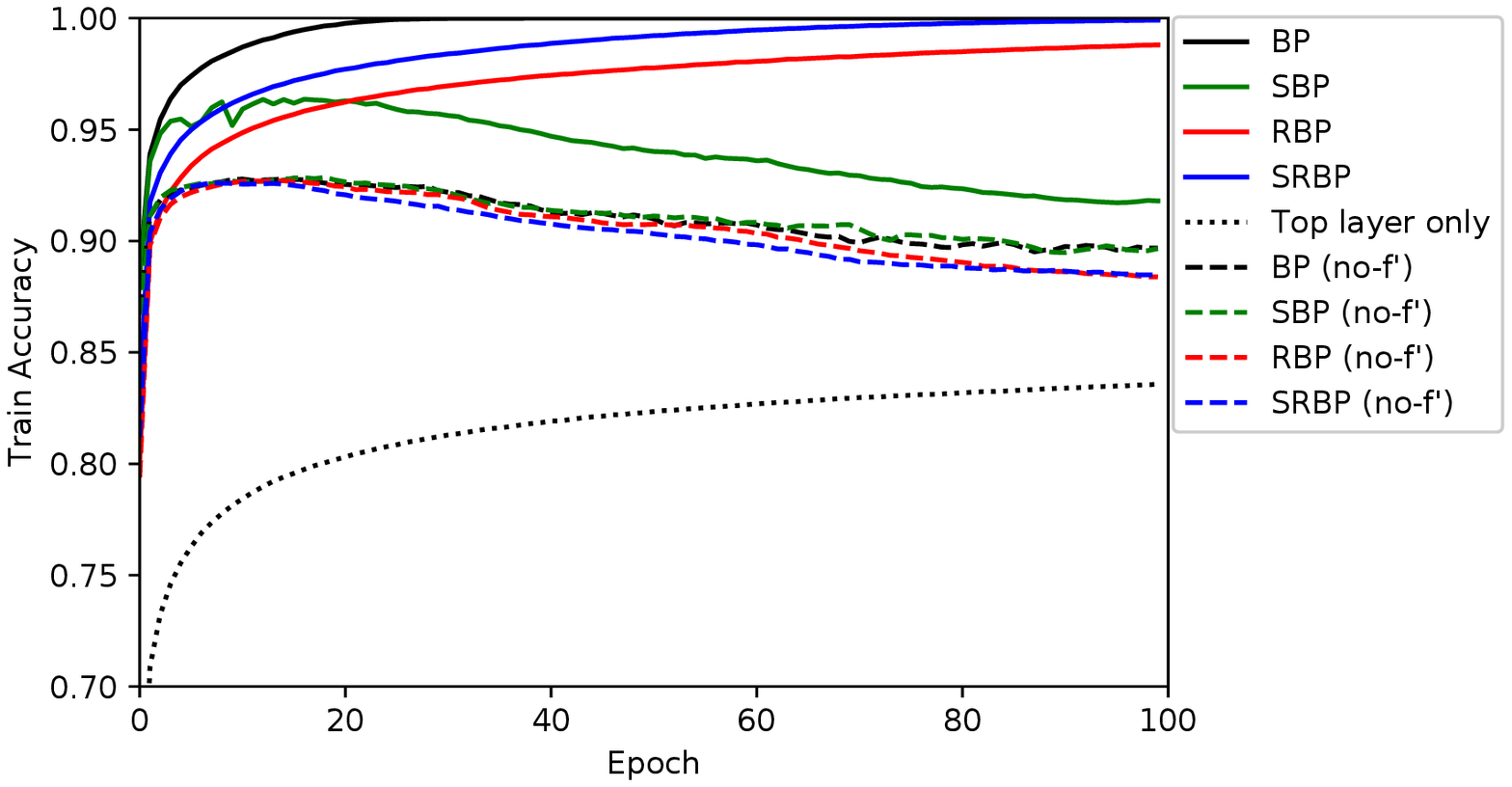}
\end{subfigure}
\begin{subfigure}[b]{\textwidth}
\includegraphics[width=\textwidth]{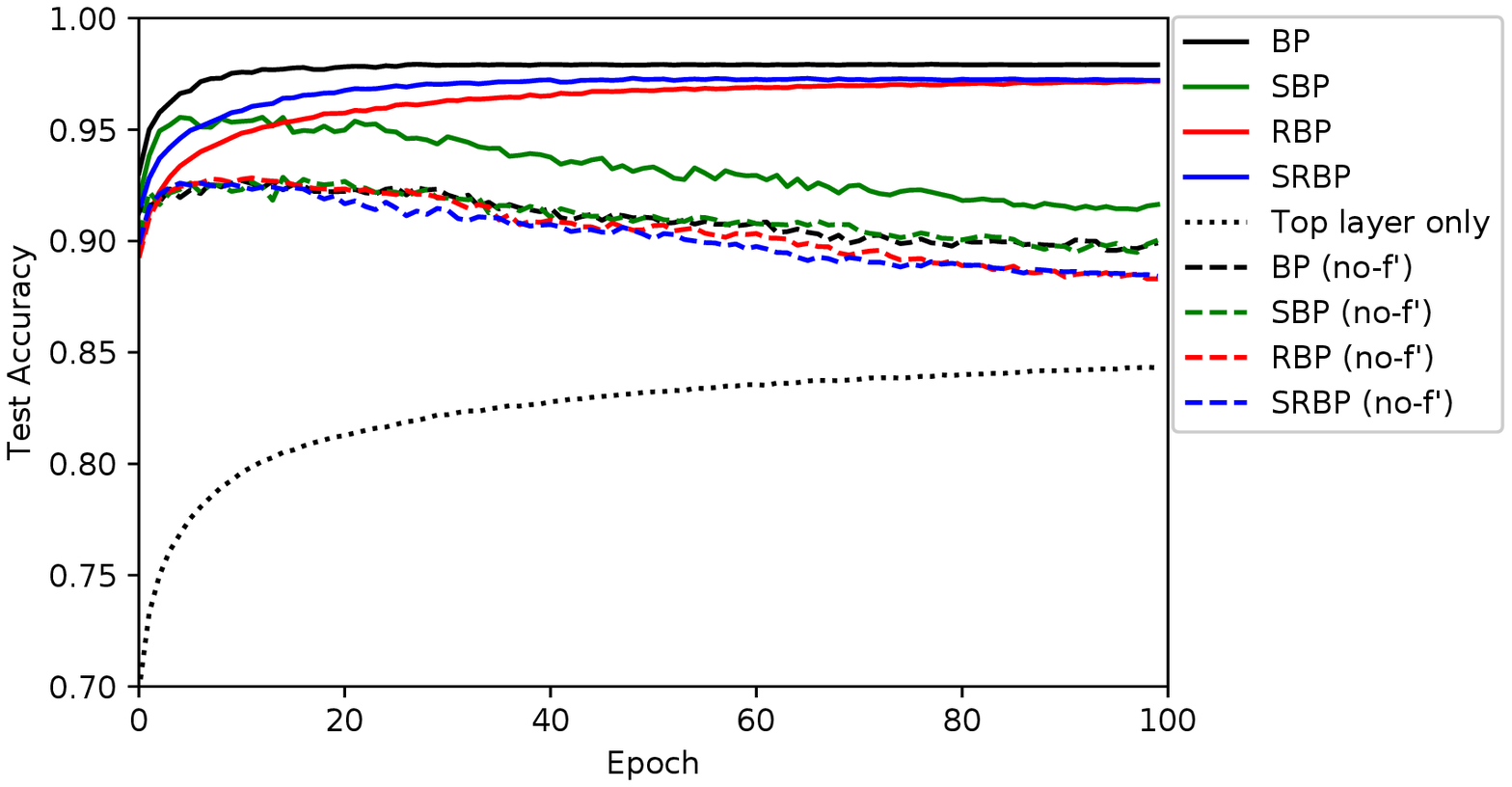}
\end{subfigure}
\caption{MNIST training (upper) and test (lower) accuracy, as a function of epoch, for nine different learning algorithms: backpropagation (BP), skip BP (SBP), random BP (RBP), skip random BP (SRBP), the version of each algorithm in which the error signal is not multiplied by the derivative of the post-synaptic transfer function (no-f'), and the case where only the top layer is trained while the lower layer weights are fixed (Top Layer Only). Note that these algorithms differ only in how they backpropagate error signals to the lower layers; the top layer is always updated according to the typical gradient descent rule. Models are trained five times with different weight initializations; the trajectory of the mean is shown here.}
\label{fig:1}
\end{figure}

\begin{figure}[H]
\begin{subfigure}[b]{\textwidth}
\includegraphics[width=\textwidth]{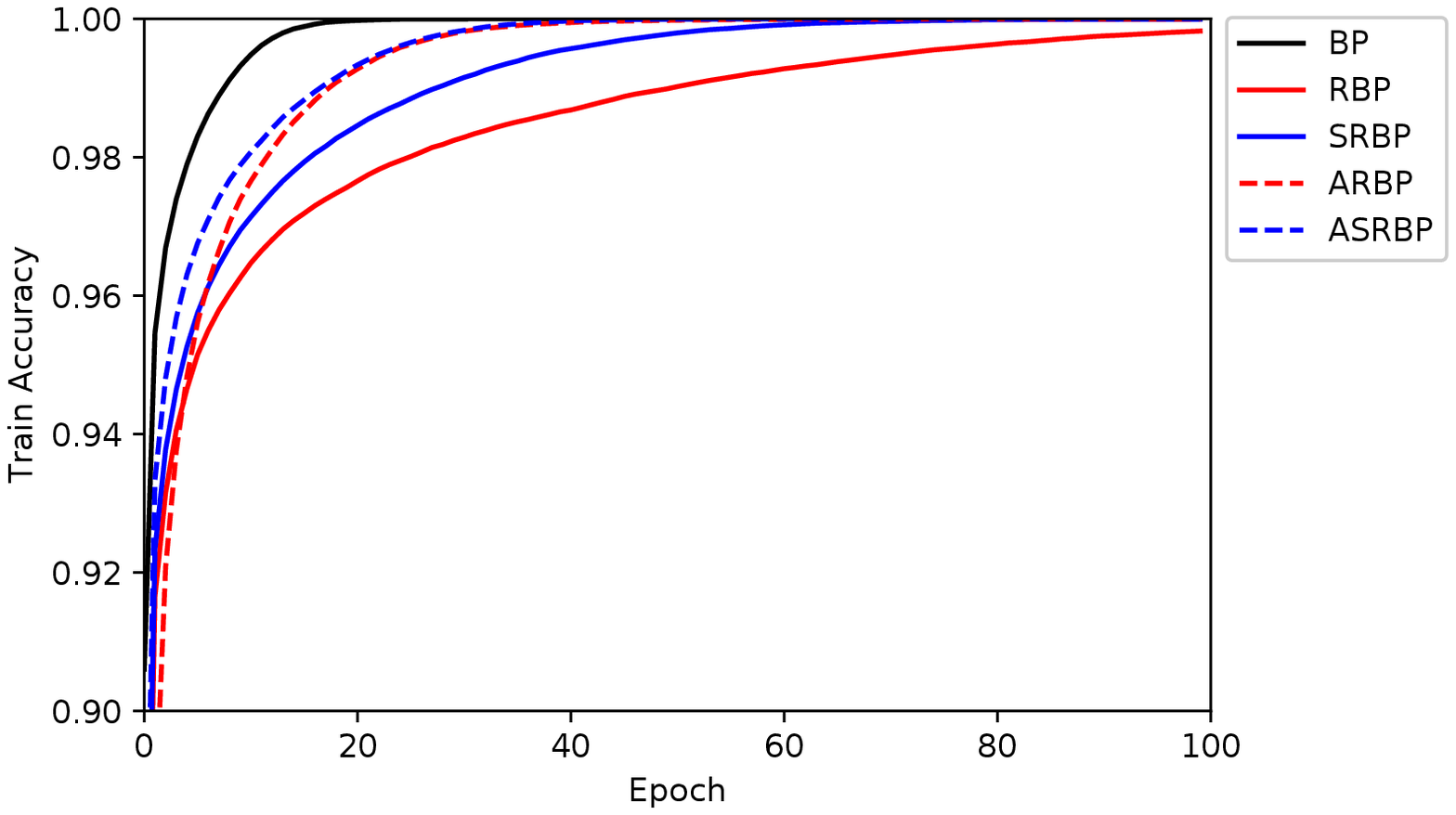}
\end{subfigure}
\begin{subfigure}[b]{\textwidth}
\includegraphics[width=\textwidth]{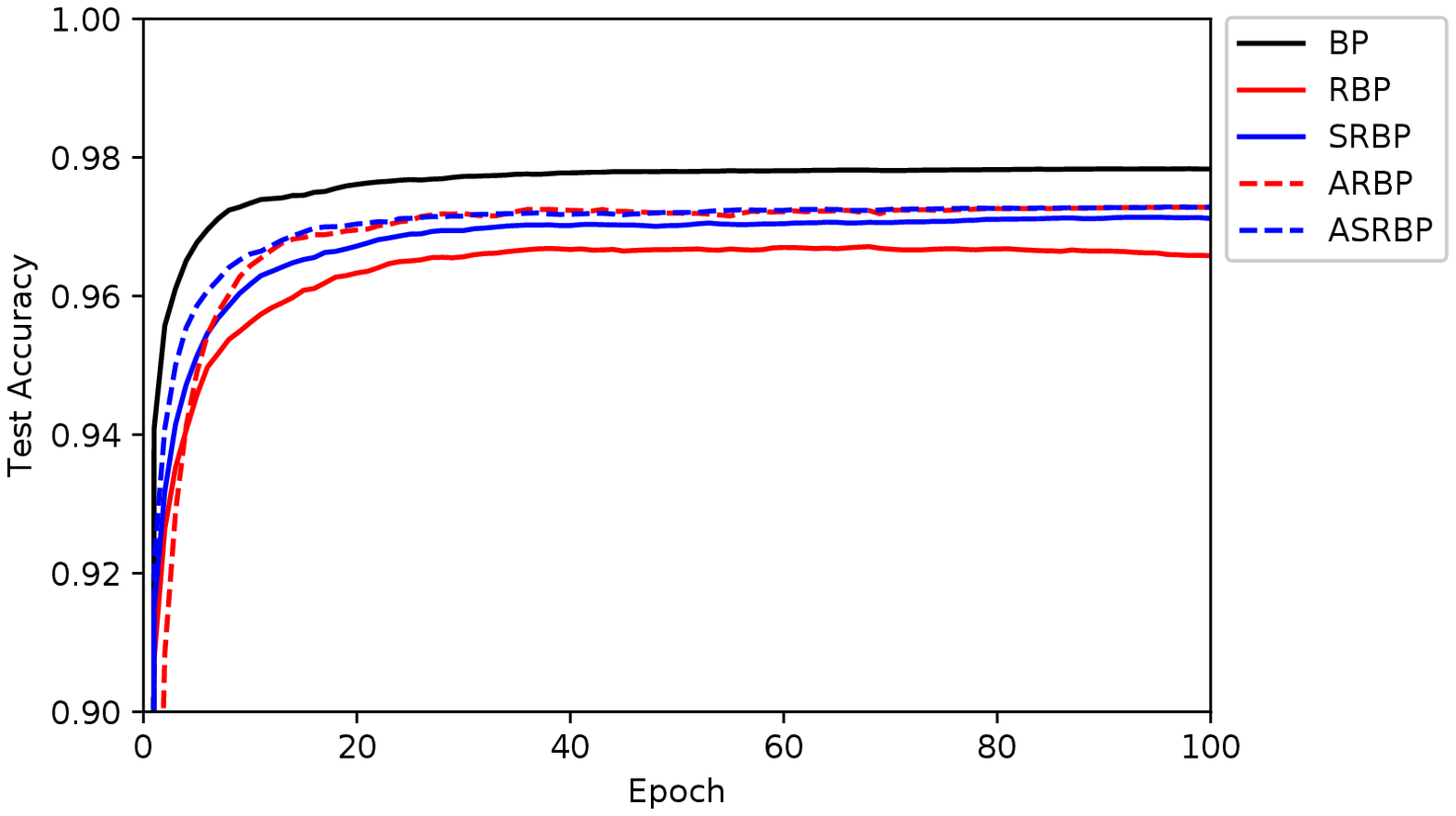}
\end{subfigure}
\caption{MNIST training (upper) and test (lower) accuracy, as a function of training epoch, for the adaptive versions of the RBP algorithm (ARBP) and SRBP algorithm (ASRBP). In these simulations, adaption slightly improves the performance of SRBP and speeds up training. For the ARBP algorithm, the learning rate was reduced by a factor of 0.1 in these experiments to keep the weights from growing too quickly. Models are trained five times with different weight initializations; the trajectory of the mean is shown here.
}
\label{fig:2}
\end{figure}

\begin{figure}[H]
\begin{subfigure}[b]{\textwidth}
\includegraphics[width=\textwidth]{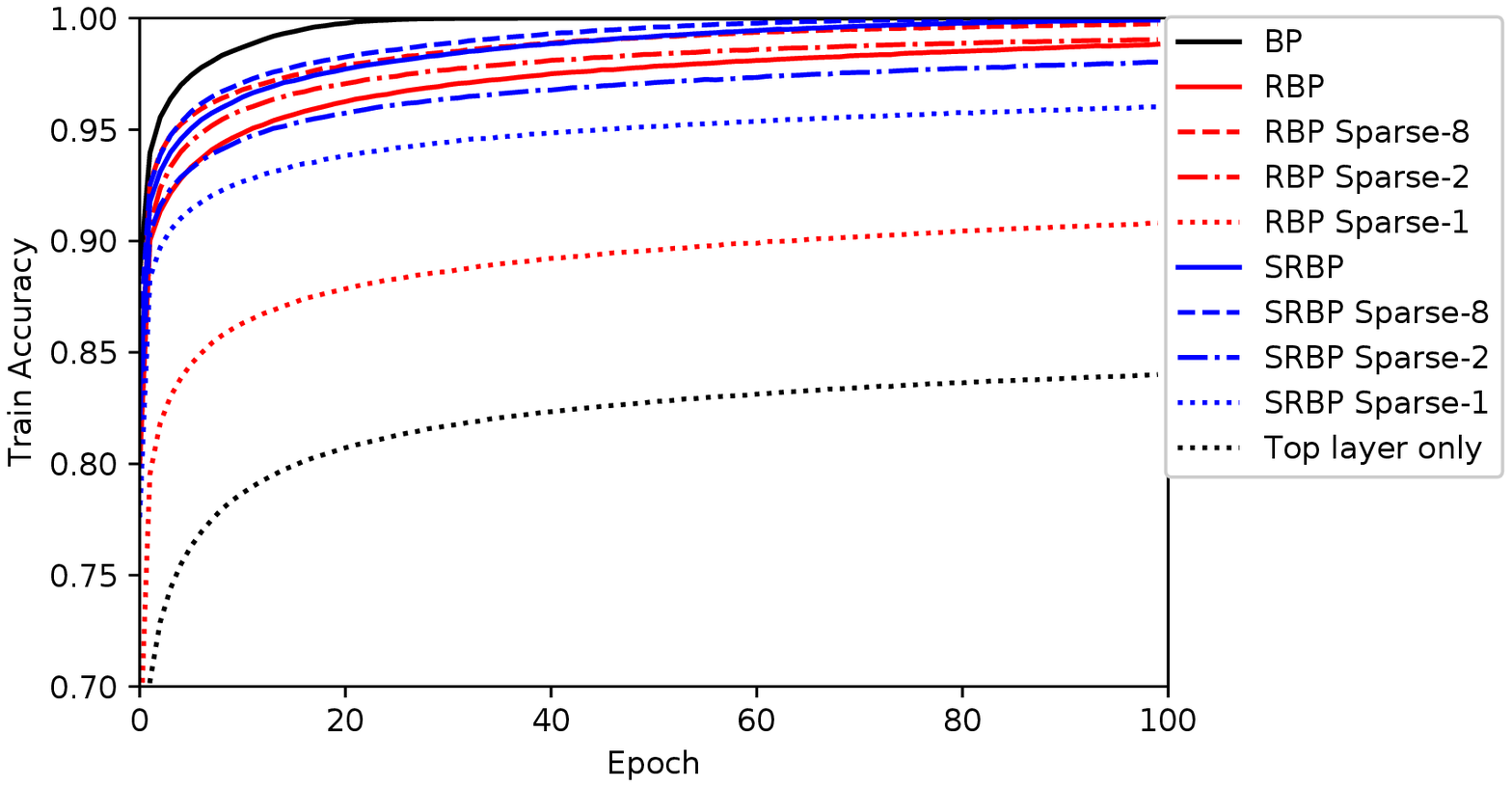}
\end{subfigure}
\begin{subfigure}[b]{\textwidth}
\includegraphics[width=\textwidth]{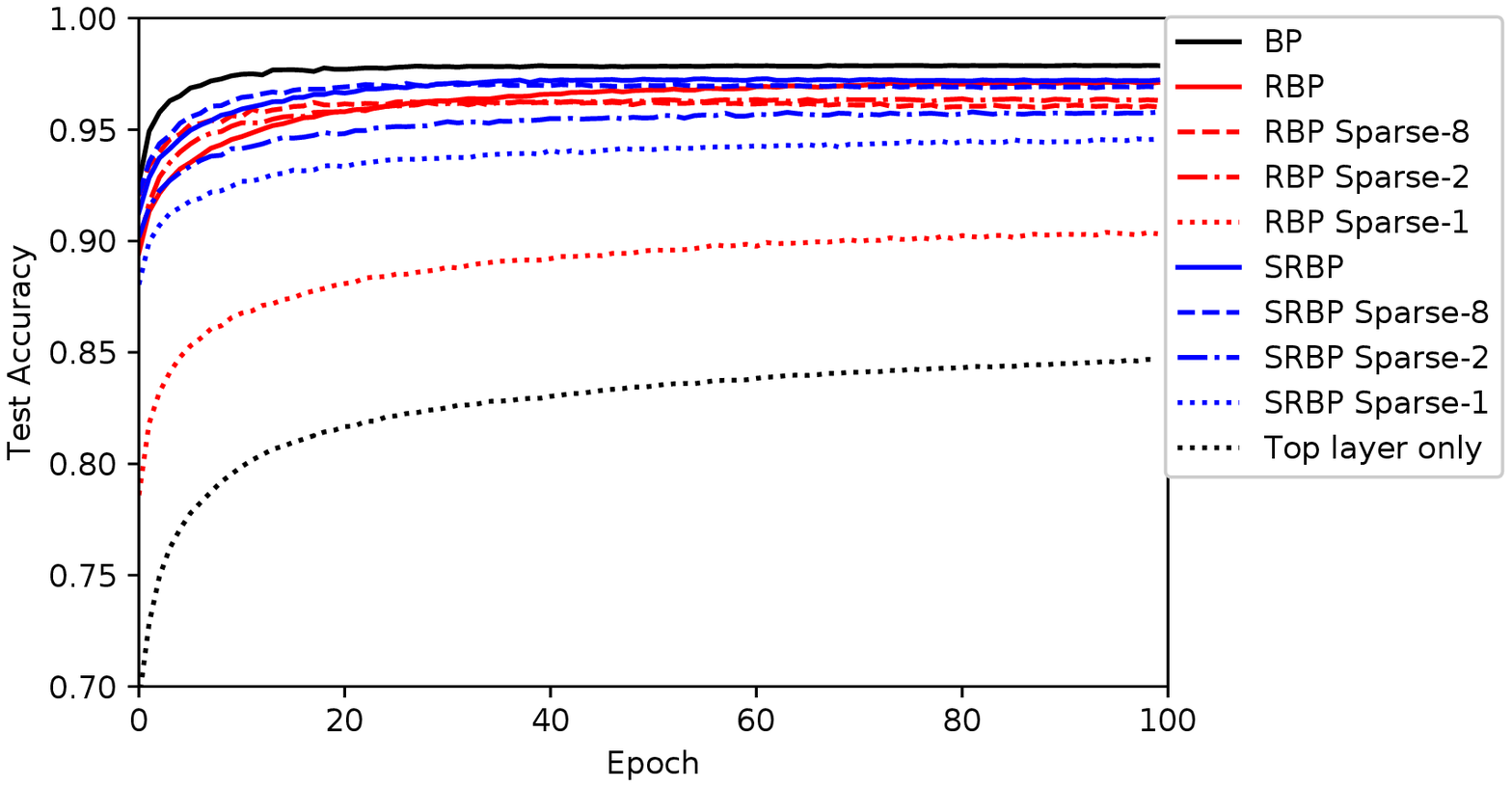}
\end{subfigure}
\caption{MNIST training (upper) and test (lower) accuracy, as a function of training epoch, for the sparse versions of the RBP and SRBP algorithms. Experiments are run with different levels of sparsity by controlling the expected number $n$ of non-zero connections sent from one neuron to any other layer it is connected to in the backward learning channel.  
The random backpropagation matrix connecting any two layers is created by sampling each entry using a (0,1) Bernoulli distribution, where each element is $1$ with probability $p={n}/{(\mathrm{fan-in})}$ and $0$ otherwise. For example, in $SRBP$ (Sparse-$1$), each of the $10$ softmax outputs sends a non-zero (hence with a weight equal to 1) connection to an average of one neuron in each of the hidden layers. We compare to the (Normal) versions of RBP and SRBP, where the elements of these matrices are initialized from a standard Normal distribution scaled in the same way as the forward weight matrices~\cite{glorot_2010}. Models are trained five times with different weight initializations; the trajectory of the mean is shown here. 
}
\label{fig:3}
\end{figure}

Several learning algorithms were first compared on the MNIST
\cite{lecun1998gradient} classification task. The neural network architecture consisted of 784 inputs, four fully-connected hidden layers of 100 tanh units, followed by 10 softmax output units. Weights were initialized by sampling from a scaled normal distribution~\cite{glorot_2010}. Training was performed for 100 epochs using mini-batches of size 100 with an initial learning rate of $0.1$, decaying by a factor of $10^{-6}$ after each update, and no momentum. In Figure \ref{fig:1}, the performance of each algorithm is shown on both the training set (60,000 examples) and test set (10,000 examples). Results for the adaptive versions of the random propagation algorithms are shown in Figure \ref{fig:2}, and results for the sparse versions are shown in Figure \ref{fig:3}.

The main conclusion is that the general concept of RBP is very robust and works almost as well as BP. Performance is unaffected or degrades gracefully when the the random backwards weights are initialized from different distributions or even change during training. The skipped versions of the algorithms seem to work slightly better than the non-skipped versions. Finally, it can be used with different neuron activation functions, though multiplying by the derivative of the activations seem to play an important role.

\begin{figure}[H]
\includegraphics[width=\textwidth]{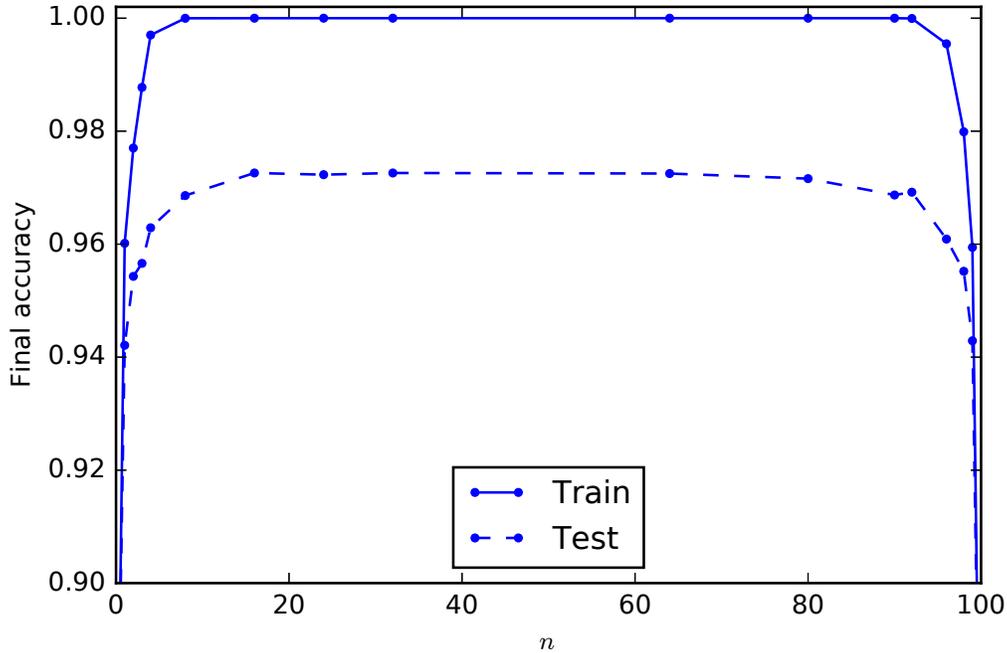}
\caption{MNIST post-training accuracy for the sparse versions of the SRBP algorithm. For extreme values of $n$, sparse SRBP fails: for $n=0$, all the backward weights are set to zero and no error signals are sent; for $n=100$ all the backward weights are set to 1, and all the neurons in a given layer receive the same error signal. The performance  of the algorithm is surprisingly robust in between these extremes. For sparse RBP (not shown), the backward weights should be scaled by a factor of $1/\sqrt{n}$ to avoid an exponential growth in the error signals of the lower layers. 
}
\label{fig:3b}
\end{figure}

\subsection{Additional MNIST Experiments}

In addition to the experiments presented above, the following observations were made by training on MNIST with other variations of these algorithms:
\begin{enumerate}
\item If the matrices of the learning channel in RBP are
randomly changed at each stochastic mini-batch update, sampled from a distribution with mean 0, performance is poor and similar to training only the top layer.
\item If the matrices of the learning channel in RBP are randomly changed at each stochastic mini-batch update, but each backwards weight is constrained to have the same sign as the corresponding forward weight, then training error goes to 0\%. This is the sign-concordance algorithm explored by Liao, et al.~\cite{liao_backpropagation_2015}. 
\item If the elements of the matrices of the learning channel in RBP or SRBP are sampled from a uniform or normal distribution with non-zero mean, performance is unchanged. This is also consistent with the sparsity experiments above, where the means of the sampling distributions are not zero.
\item Updates to a deep layer with RBP or SRBP appear to require updates in the precedent layers in the learning channel. If we fix the weights in layer $h$, while updating the rest of the layers with SRBP, performance is often worse than if we fix layers $l \leq h$.
\item If we remove the magnitude information from the SRBP updates, keeping only the sign, performance is better than the Top Layer Only algorithm, but not as good as SRBP. This is further explored in the next section.
\item If we remove the sign information from the SRBP updates, keeping only the absolute value, things do not work at all.
\item If a different random backward weight is used to send an error signal to each individual \emph{weight}, rather  than to a hidden neuron which then updates all it's incoming weights, things do not work at all.
\item The RBP learning rules work with different transfer functions as well, including linear, logistic, and ReLU (rectified linear) units.
\end{enumerate}

\subsection{CIFAR-10}

To further test the validity of these results, we performed similar simulations with a convolutional architecture on the CIFAR-10 dataset \cite{krizhevsky2009learning}. The specific architecture was based on previous work \cite{hinton_improving_2012}, and consisted of 3 sets of convolution and max-pooling layers, followed by a densely-connected layer of 1024 tanh units, then a softmax output layer. The input consists of 32-by-32 pixel 3-channel images; each convolution layer consists of 64 tanh channels with 5x5 kernel shape and 1x1 strides; max-pooling layers have 3x3 receptive fields and 2x2 strides. All weights were initialized by sampling from a scaled normal distribution~\cite{glorot_2010}, and updated using stochastic gradient descent on mini-batches of size 128 and a momentum of $0.9$. The learning rate started at $0.01$ and decreased by a factor of $10^{-5}$ after each update. During training, the training images are randomly translated up to 10\% in either direction, horizontally and vertically, and flipped horizontally with probability $p=0.5$.

\begin{figure}[H]
\begin{subfigure}[b]{\textwidth}
\includegraphics[width=\textwidth]{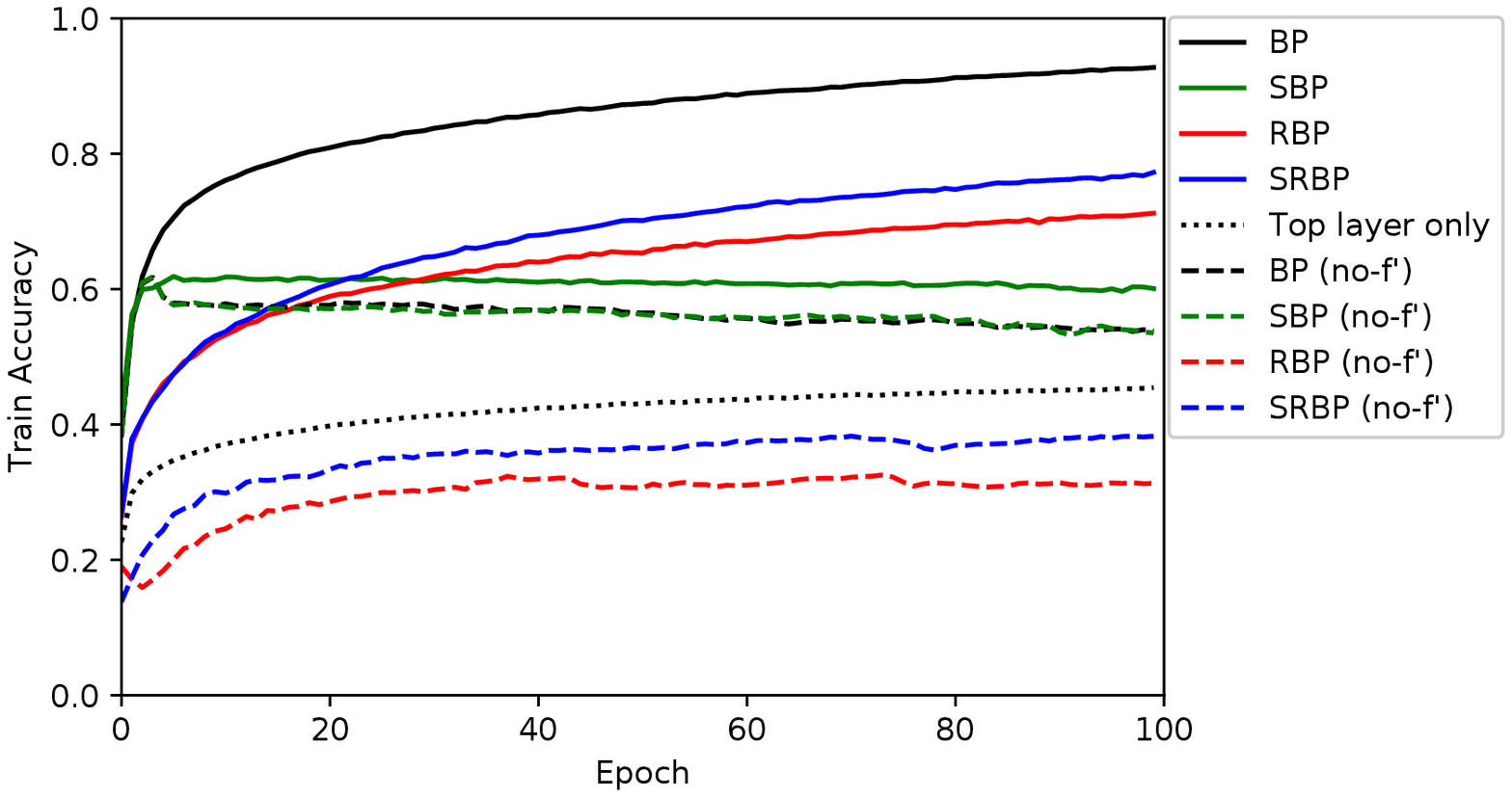}
\end{subfigure}
\begin{subfigure}[b]{\textwidth}
\includegraphics[width=\textwidth]{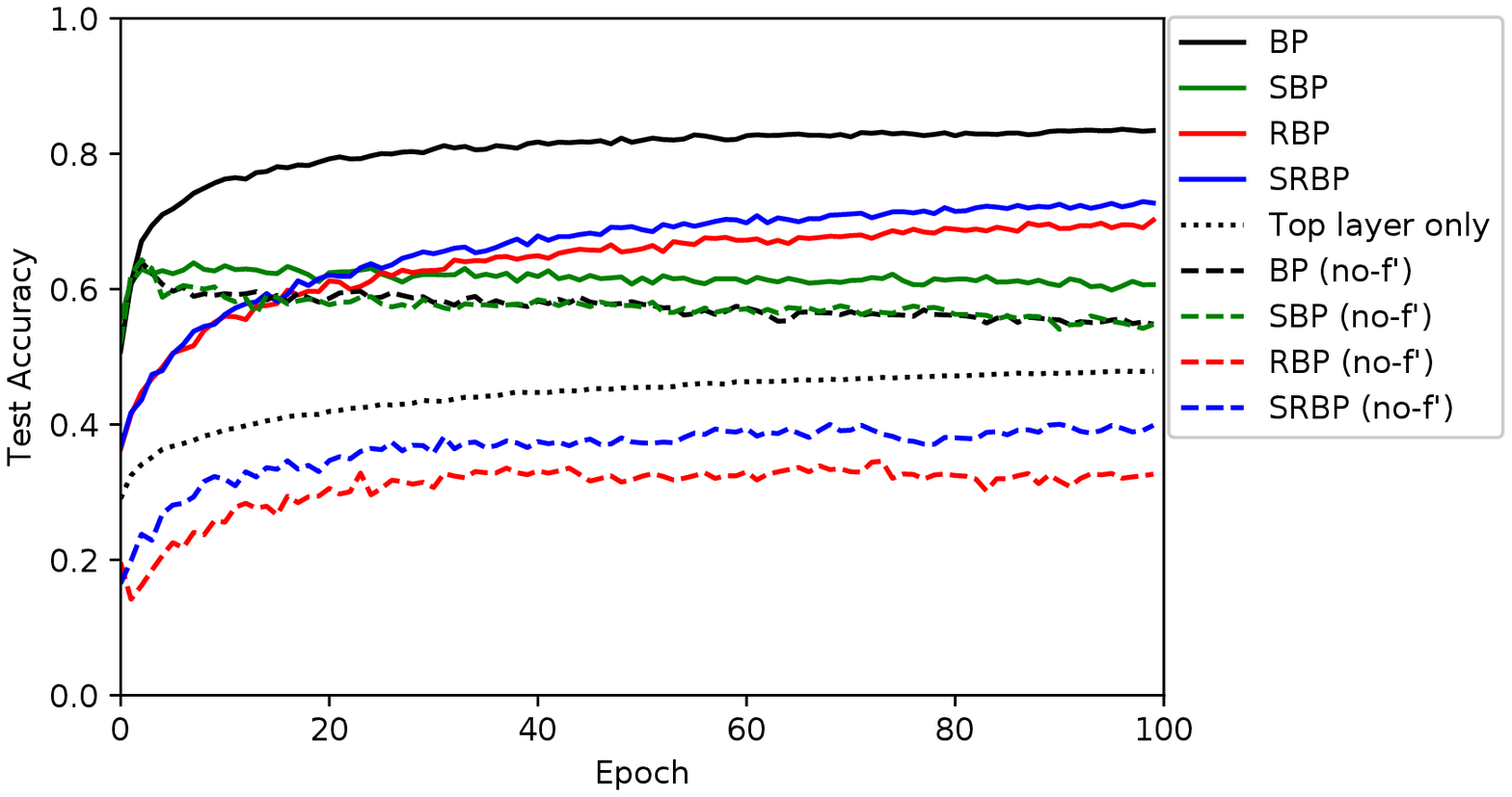}
\end{subfigure}
\caption{CIFAR-10 training (upper) and test (lower) accuracy, as a function of training epoch, for nine different learning algorithms: backpropagation (BP), skip BP (SBP), random BP (RBP), skip random BP (SRBP), the version of each algorithm in which the error signal is not multiplied by the derivative of the post-synaptic transfer function (no-f'), and the case where only the top layer is trained while the lower layer weights are fixed (Top Layer Only). Models are trained five times with different weight initializations; the trajectory of the mean is shown here.}
\label{fig:4}
\end{figure}

\begin{figure}[H]
\begin{subfigure}[b]{\textwidth}
\includegraphics[width=\textwidth]{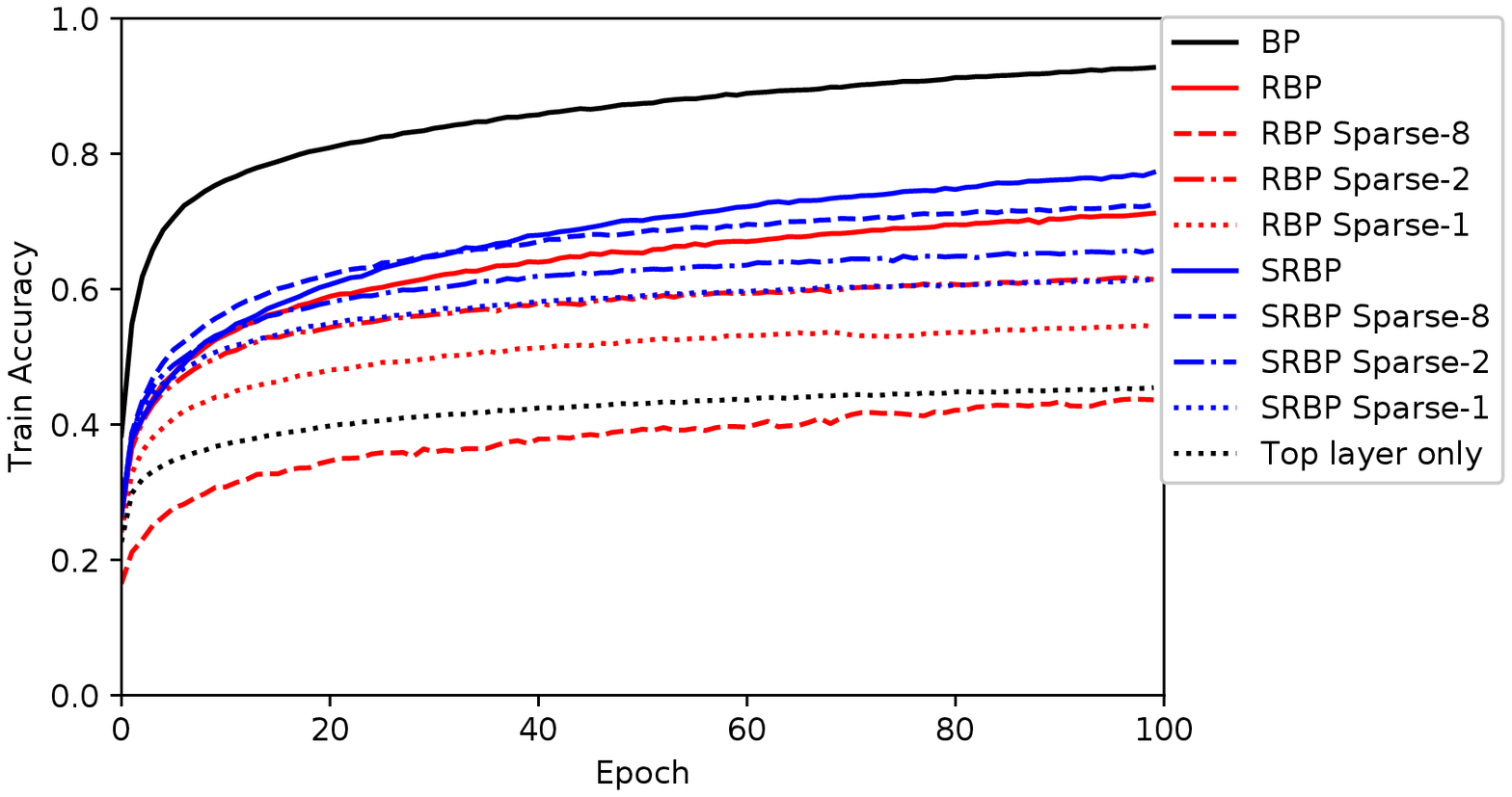}
\end{subfigure}
\begin{subfigure}[b]{\textwidth}
\includegraphics[width=\textwidth]{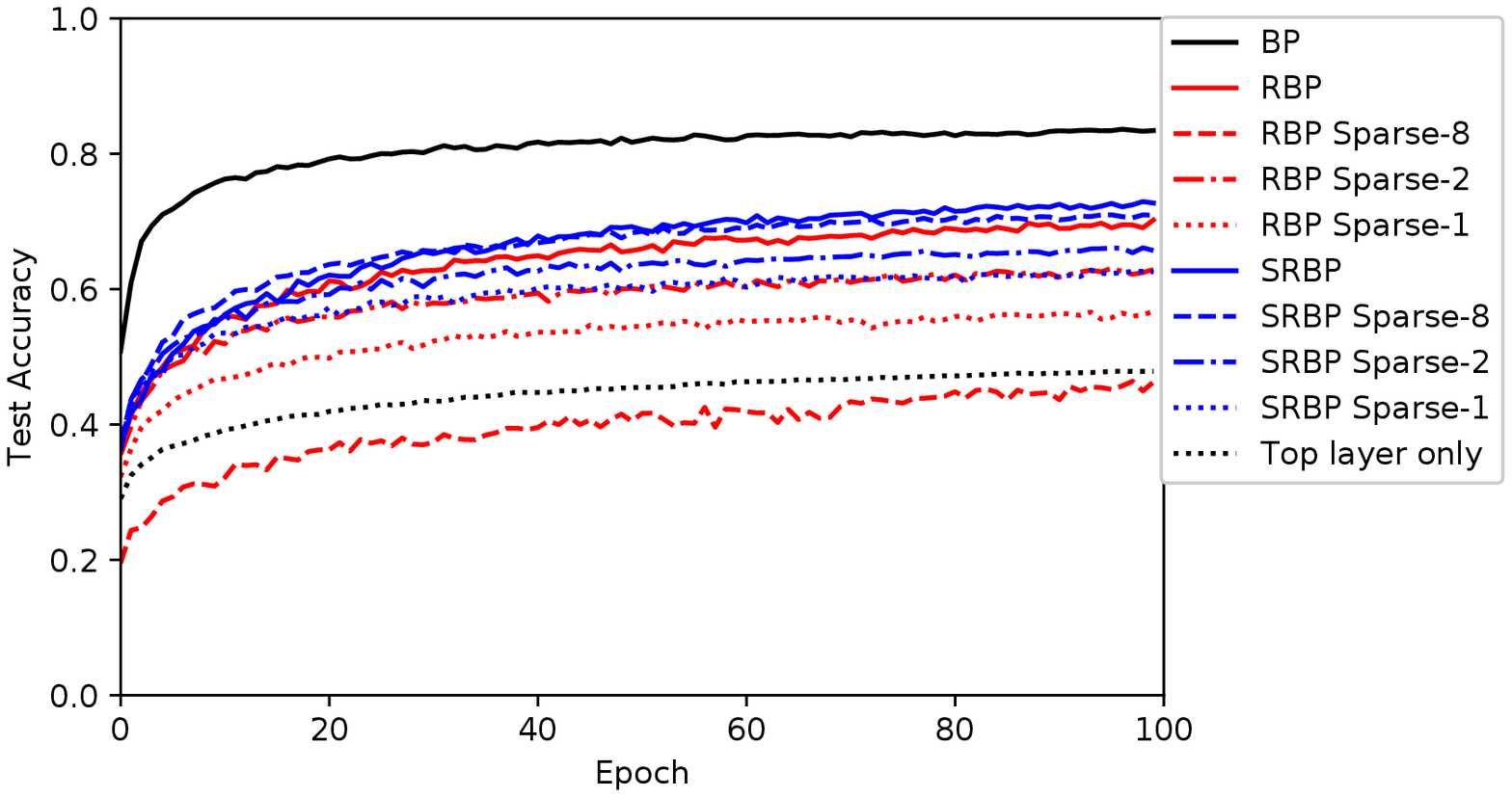}
\end{subfigure}
\caption{
CIFAR-10 training (upper) and test (lower) accuracy for the sparse versions of the RBP and SRBP algorithms. 
Experiments are run with different levels of sparsity by controlling the expected number $n$ of non-zero connections sent from one neuron to any other layer it is connected to in the backward learning channel.  
The random backpropagation matrix connecting any two layers is created by sampling each entry using a (0,1) Bernoulli distribution, where each element is $1$ with probability $p={n}/{(\mathrm{fan-in})}$ and $0$ otherwise. We compare to the (Normal) versions of RBP and SRBP, where the elements of these matrices are initialized from a standard Normal distribution scaled in the same way as the forward weight matrices~\cite{glorot_2010}. Models are trained five times with different weight initializations; the trajectory of the mean is shown here.
}
\label{fig:5}
\end{figure}

Examples of results obtained with these 2D convolutional architectures are shown in Figures \ref{fig:4}
and \ref{fig:5}. Overall they are very similar to those obtained on the MNIST dataset.

\begin{table}[H]
\begin{center}
\begin{tabular}{ | l | c c c c| } 
\hline
 & BP & RBP & SRBP & Top layer only\\
\hline\hline
 MNIST Baseline      & 97.9 (0.1) & 97.2 (0.1) & 97.2 (0.2) & 84.7 (0.7)\\ 
\hline
No-$f^\prime$ & 89.9 (0.3) &  88.3 (1.1) & 88.4 (0.7) & \\
\hline 
Adaptive & & 97.3 (0.1) & 97.3 (0.1) &\\ 
\hline
Sparse-8      && 96.0 (0.4) & 96.9 (0.1) &\\ 
Sparse-2      && 96.3 (0.5) & 95.8 (0.2) &\\
Sparse-1      && 90.3 (1.1) & 94.6 (0.6) &\\
\hline
Quantized error 5-bit & 97.6 & 95.4 & 95.1 &\\
Quantized error 3-bit & 96.5 & 92.5 & 93.2 &\\
Quantized error 1-bit & 94.6 & 89.8 & 91.6 &\\ 
\hline
Quantized update 5-bit & 95.2 & 94.0 & 93.3 &\\ 
Quantized update 3-bit & 96.5 & 91.0 & 92.2 &\\ 
Quantized update 1-bit & 92.5 & 9.6 & 90.7 &\\ 
\hline
LC Dropout 10\% & 97.7 & 96.5  & 97.1 &\\ 
LC Dropout 20\% & 97.8 & 96.7  & 97.2 &\\ 
LC Dropout 50\% & 97.7 & 96.7  & 97.1 &\\ 
\hline\hline
CIFAR-10 Baseline & 83.4 (0.2) & 70.2 (1.1) & 72.7 (0.8) & 47.9 (0.4) \\
\hline
No-$f^\prime$ & 54.8 (3.6)  & 32.7 (6.2) & 39.9 (3.9) &\\
\hline
Sparse-8      && 46.3 (4.3)  & 70.9 (0.7) &\\ 
Sparse-2      && 62.9 (0.9)  & 65.7 (1.9) &\\
Sparse-1      && 56.7 (2.6) & 62.6 (1.8) &\\
\hline
\end{tabular}
\end{center}
\caption{Summary of experimental results showing the final test accuracy (in percentages) for the RBP algorithms after 100 epochs of training on MNIST and CIFAR-10. For the experiments in this section, training was repeated five times with different weight initializations; in these cases the mean is provided, with the sample standard deviation in parentheses. Also included are the quantization results from Section 5, and the experiments applying dropout to the learning channel from Section 6.}
\label{tab:summary}
\end{table}

\section{Bit Precision in the Learning Channel}

\subsection{Low-Precision Error Signals}

In the following experiment, we investigate the nature of the learning channel by quantizing the error signals in the BP, RBP, and SRBP algorithms. This is distinct from other work that uses quantization to reduce computation~\cite{hubara_quantized_2016} or memory~\cite{han_compression_2015} costs. Quantization is not applied to the forward activations or weights; quantization is only applied to the backpropagated signal received by each hidden neuron, $J^h_i(T-O^L)$, where each weight update after quantization is given by

\begin{eqnarray}
\Delta w^h_{ij} &=& I^h_i(T-O^L) \times O^{h-1}_j \\
 &=& \mathrm{Quantize} \left( J^h_i(T-O^L) \right) \times (f^h_i)^\prime \times O^{h-1}_j
\end{eqnarray}
where $(f^h_i)^\prime$ is the derivative of the activation function and

\begin{eqnarray}
I^h_i(T-O^L) &= & J^h_i(T-O^L)  \times (f^h_i)^\prime
\end{eqnarray}
in the non-quantized update. We define the quantization formula used here as

\begin{eqnarray}
\label{eq:quantize}
\mathrm{Quantize}_{\alpha, bits} (x) = \alpha \times \mathrm{sign}(x) \times 2^{\mathrm{round}\left( \mathrm{clip}(\mathrm{log2}\vert \frac{x}{\alpha} \vert,\ -bits+1,\ 0)\right)}
\label{eq:quantize}
\end{eqnarray}
where $bits$ is the number of bits needed to represent $2^{bits}$ possible values and $\alpha$ is a scale factor such that the quantized values fall in the range $[-\alpha, \alpha]$. Note that this definition is identical to the quantization function defined in Hubara, et al.~\cite{hubara_quantized_2016}, except that this definition is more general in that $\alpha$ is not constrained to be a power of 2.

\begin{figure}[H]
\begin{subfigure}[b]{\textwidth}
\includegraphics[width=\textwidth]{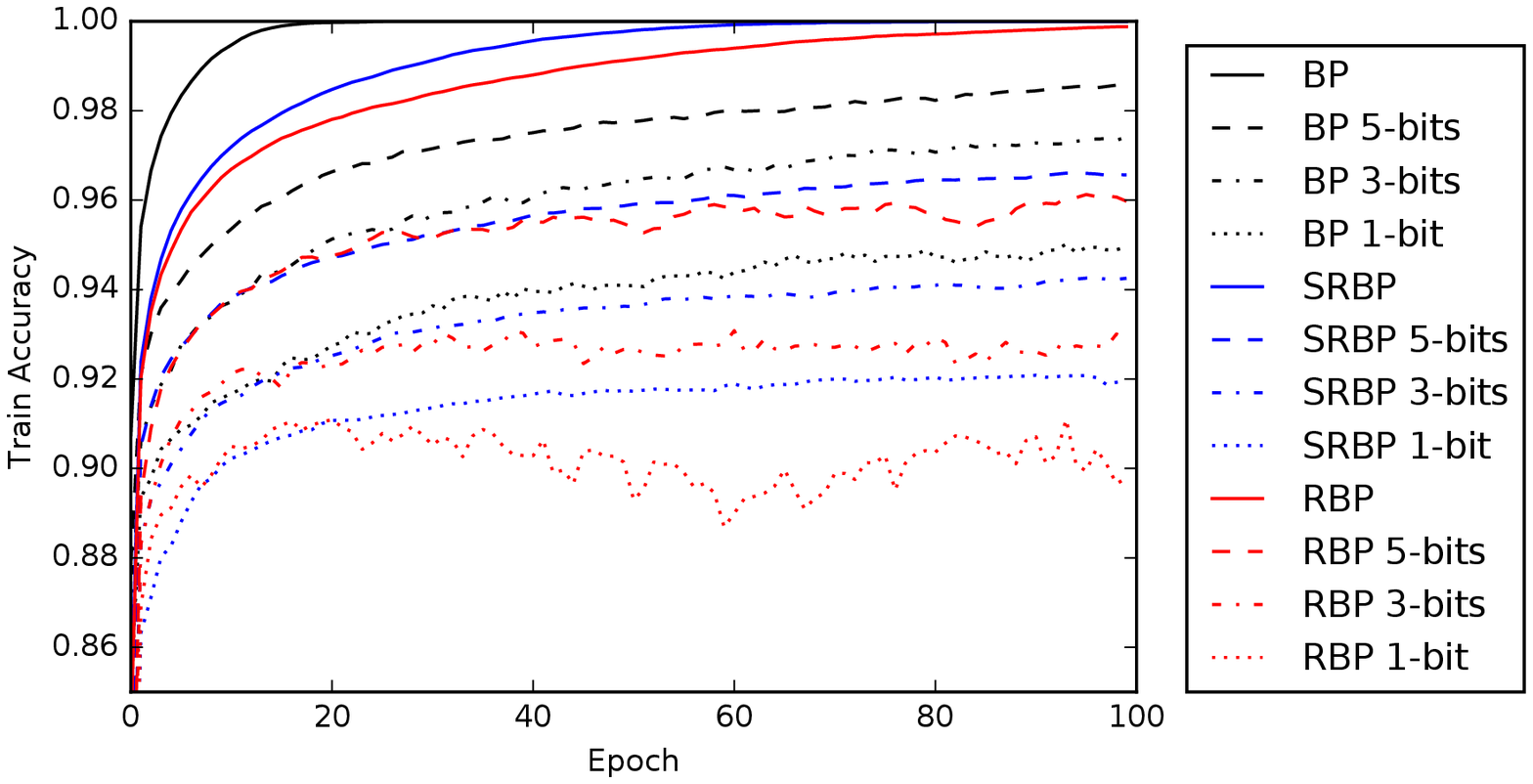}
\end{subfigure}
\begin{subfigure}[b]{\textwidth}
\includegraphics[width=\textwidth]{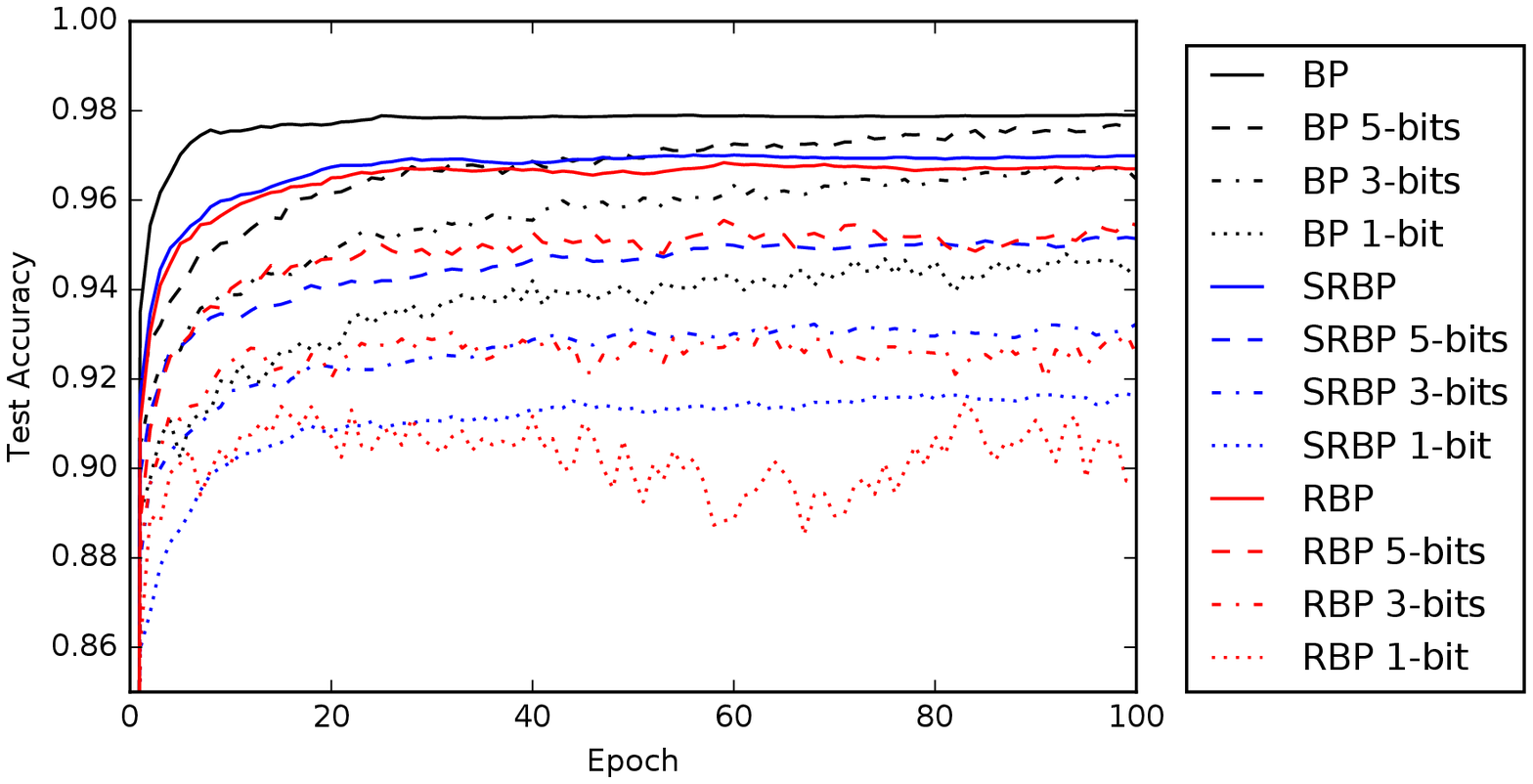}
\end{subfigure}
\caption{MNIST training (upper) and test (lower) accuracy, as a function of training epoch, for the sparse versions of the RBP and SRBP algorithms. Experiments are run with different levels of quantization of the error signal by controlling the bitwidth $bits$, according to the formula given in the text (Equation \ref{eq:quantize}).} 
\label{fig:quantize_errors_mnist}
\end{figure}

In BP and RBP, the quantization occurs \emph{before} the error signal is backpropagated to previous layers, so the quantization errors accumulate. In experiments, we used a fixed scale parameter $\alpha=2^{-3}$ and varied the bit width $bits$. Figure \ref{fig:quantize_errors_mnist} shows that the performance degrades gracefully as the precision of the error signal decreases to small values; for larger values, e.g. $bits=10$, the performance is indistinguishable from the unquantized updates with 32-bit floats.

\subsection{Low-Precision Weight Updates}

\begin{figure}[H]
\begin{subfigure}[b]{\textwidth}
\includegraphics[width=\textwidth]{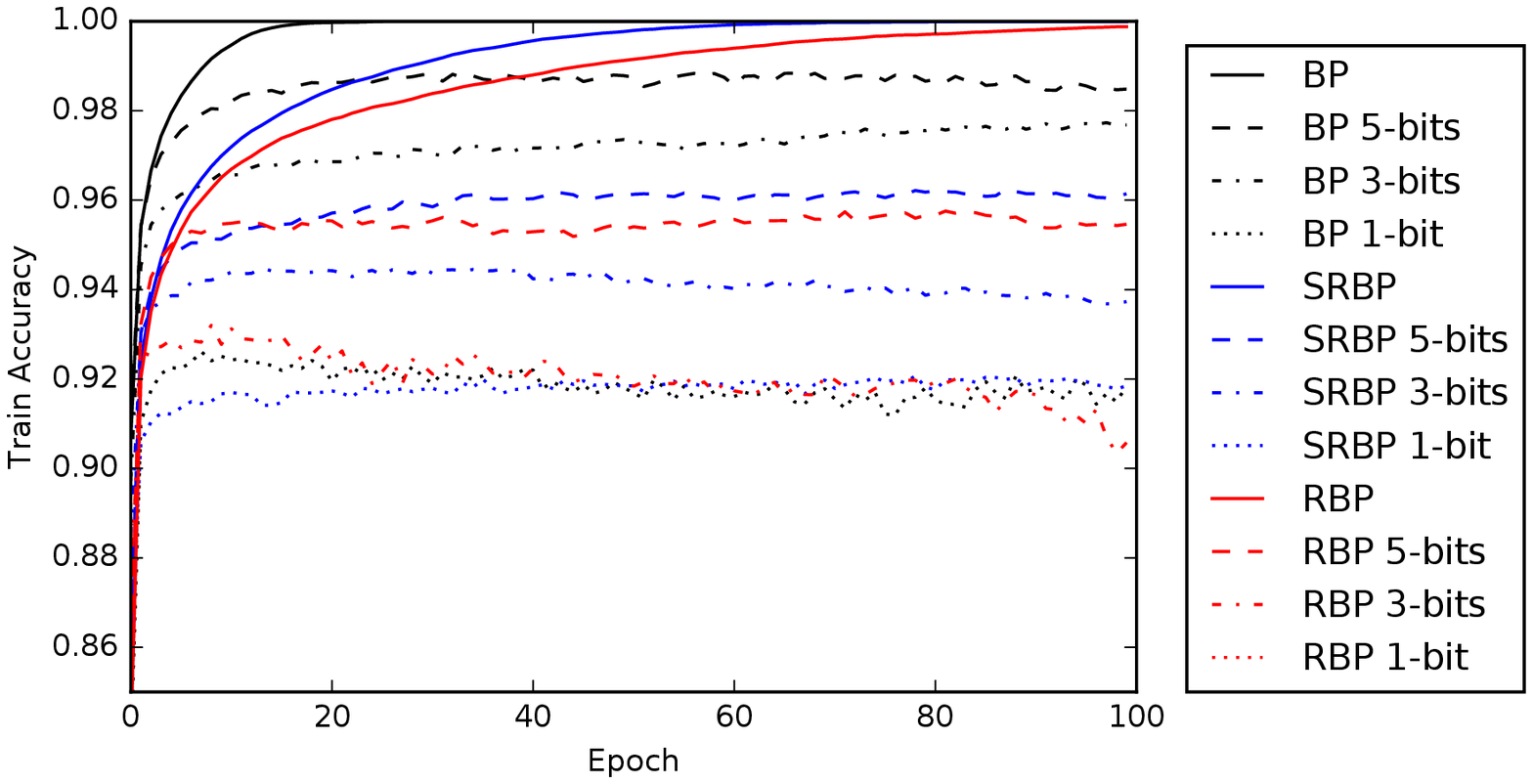}
\end{subfigure}
\begin{subfigure}[b]{\textwidth}
\includegraphics[width=\textwidth]{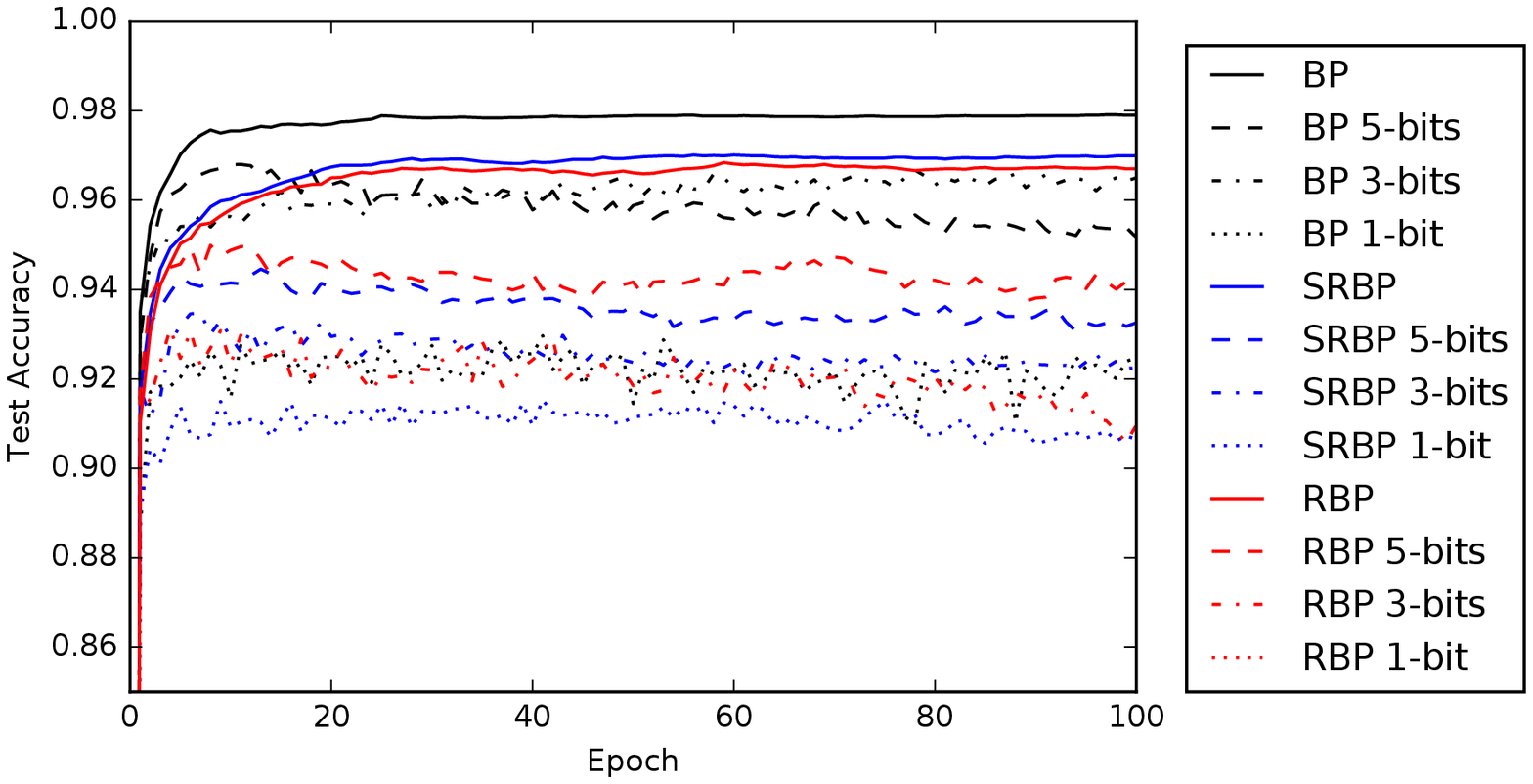}
\end{subfigure}
\caption{MNIST training (upper) and test (lower) accuracy, as a function of training epoch, for the sparse versions of the RBP and SRBP algorithms. Experiment are carried with different levels of quantization of the weight updates by controlling the bitwidth $bits$, according to the formula given in the text (Equation \ref{eq:quantize}). Quantization is applied to each example-specific update, before summing the updates within a minibatch.}
\label{fig:quantize_updates_mnist}
\end{figure}

The idea of using low-precision weight updates is not new~\cite{riedmiller_rprop_1993}, and Liao, et al.~\cite{liao_backpropagation_2015} recently explored the use of low-precision updates with RBP. In the following experiment, we investigate the robustness of both RBP and SRBP to low-precision weight updates by controlling the degree of quantization. Equation \ref{eq:quantize} is again used for quantization, with the scale factor reduced to $\alpha=2^{-6}$ since weight updates need to be small. The quantization is applied after the error signals have been backpropagated to all the hidden layers, but before summing over the minibatch; as in the previous experiments, we use minibatch updates of size 100, a non-decaying learning rate of 0.1, and no momentum term
(Figure \ref{fig:quantize_updates_mnist}). The main conclusion is that even very low-precision updates to the weights can be used to train an MNIST classifier to 90\% accuracy, and that low-precision weight updates appear to degrade the performance of BP, RBP, and SRBP in roughly the same way.

\section{Observations}

In this section, we provide a number of simple observations that provide some intuition for some of the previous simulation results and why RBP and some of its variations may work. Some of these observations are focused on SRBP which in general is easier to study than standard RBP.

\noindent
{\bf Fact 1: } In all these RBPs algorithms, the $L$-layer at the top with parameters 
$w_{ij}^L$ follows the gradient, as it is trained just like BP, since there are no random feedback weights used for learning in the top layer. In other words, BP=RBP=SRBP for the top layer.

\noindent
{\bf Fact 2: } For a given input, if the sign of $T-O$ is changed, all the weights updates are changed in the opposite direction. This is true of all the algorithms considered here--BP, RBP, and their variants--even when the derivatives of the activations are included. 

\noindent
{\bf Fact 3:} In all RBP algorithms, if $T-O=0$ (online or in batch mode) then 
for all the weights $\Delta w_{ij}^h=0$ (on line or in batch mode).

\noindent
{\bf Fact 4:} Congruence of weights is not necessary. However it can be helpful sometimes and speed up learning. 
This can easily be seen in simple cases. For instance, 
consider a linear or non-linear 
${\cal A}[N_0,N_1,1]$ architecture with coherent weights, and denote by $a$ the weights in the bottom layer, by $b$ the weights in the top layer, and by $c$ the weights in the learning channel. Then, for all variants of RBP, all the weights updates are in the same direction as the gradient. This is obvious for the top layer (Fact 1 above). For the first layer of weights, the changes are given by $\Delta w^1_{ij}=\eta (T-O)c_iI_j$, which is very similar to the change produced by gradient descent $\Delta^1_{ij}=\eta (T-O)b_iI_j$ since $c_i$ and $b_i$ are assumed to be coherent. So while the dynamics of the lower layer is not exactly in the gradient direction, it is always in the  same orthant as the gradient and thus downhill with respect to the error function. Additional examples showing the positive but not necessary effect of coherence are given in Section 7.

\noindent
{\bf Fact 5:} SRBP seems to perform well showing that the upper derivatives are not needed. However the derivative of the corresponding layer seem to matter. In general, for the activation functions considered here, these derivatives tend to be between 0 and 1. Thus learning is attenuated for neurons that are saturated. So an ingredient that seems to matter is to let the synapses of neurons that are not saturated change more than the synapses of neurons that are saturated ($f'$ close to 0).

\noindent
{\bf Fact 6:} Consider a multi-class classification problem, such as MNIST. All the elements in the same class tend to receive the same backpropagated signal and tend to move in unison. For instance, consider the the beginning of learning, with small random weights in the forward network.
Then all the images will tend to produce a more or less uniform output vector similar to $(0.1,0.1, \ldots, 0.1)$.
Thus all the images in the ``0'' class will tend to produce a more or less uniform error vector similar to $(0.9, -0.1,\ldots ,-0.1)$. All the images in the ``1'' class will 
tend to produce a more or less uniform error vector similar to $(-0.1,0.9, \ldots ,-0.1)$, which is essentially orthogonal to the previous error vector, and so forth. In other words, the 10 classes can be associated with 10 roughly orthogonal error vectors. When these vectors are multiplied by a fixed random matrix, as in SRBP, they will tend to produce 10 approximately orthogonal vectors in the corresponding hidden layer. Thus the backpropagated error signals tend to be similar within one digit class, and orthogonal across different digit classes. At the beginning of learning, we can expect roughly half of them (5 digits out of 10 in the MNIST case) to be in the same direction as BP.

Thus, in conclusion, an intuitive picture of why RBP may work is that: (1) the random weights introduce a fixed coupling between the learning dynamics of the forward weights (see also mathematical analyses below); (2) the top layer of weights always follows gradient descent and stirs the learning dynamic in the right direction; and (3) the learning dynamic tends to cluster inputs associated with the same response and move them away from other similar clusters. Next we discuss a possible connection to dropout.

\subsection{Connections to Dropout}

Dropout \cite{hinton_improving_2012,baldidropout14} is a very different training algorithm which, however, is also based on using some form of randomness. 
Here we explore some possible connections to RBP.

First observe that the BP equations can be viewed as a form of dropout averaging equations, in the sense that, for a fixed example, they compute the ensemble average activity of all the units in the learning channel. The ensemble average is taken over all the possible backpropagation networks where each unit is dropped stochastically, unit $i$ in layer $h$ being dropped with probability $1-f'(S_o^h)$ [
assuming the derivatives of the transfer functions are always between 0 and 1 inclusively, which is the case for the standard transfer functions, such as the logistic or the rectified linear transfer functions--otherwise some rescaling is necessary]. Note that in this way the dropout probabilities change with each example and units that are more saturated are more likely to be dropped, consistently with the remark above that saturated units should learn less. 

In this view there are two kinds of noise: (1) choice of the dropout probabilities which vary with each example; (2) the actual dropout procedure. Consider now adding a third type of noise on all the symmetric weights in the backward pass in the form

\be
w_{ij}^h+ \xi_{ij}^h
\label{eq:dropout1}
\ee
and assume for now that $E(\xi_{ij}^h)=0$. The distribution of the noise could be Gaussian for instance, but this is not essential. The important point is that the noise on a weight is independent of the noise on the other weights, as well as independent of the dropout noise on the  units.
Under these assumptions, as shown in \cite{baldidropout14}, the expected value of the activity of each unit in the backward pass is exactly given by the standard BP equations and equal to $B_i^h $ for unit $i$ in layer $h$. {\it In other words, standard backpropagation can be viewed as computing the exact average over all backpropagation processes implemented on all the stochastic realizations of the backward network under the three forms of noise described above.} Thus we can reverse this argument and consider  that RBP approximates this average or BP by
averaging over the first two kinds of noise, but not the third one where, instead of averaging, a random realization of the weights is selected and then fixed at all epochs. This connection suggests other intermediate RBP variants where several samples of the weights are used, rather than a single one.

\begin{figure}[H]
\begin{subfigure}[b]{\textwidth}
\includegraphics[width=\textwidth]{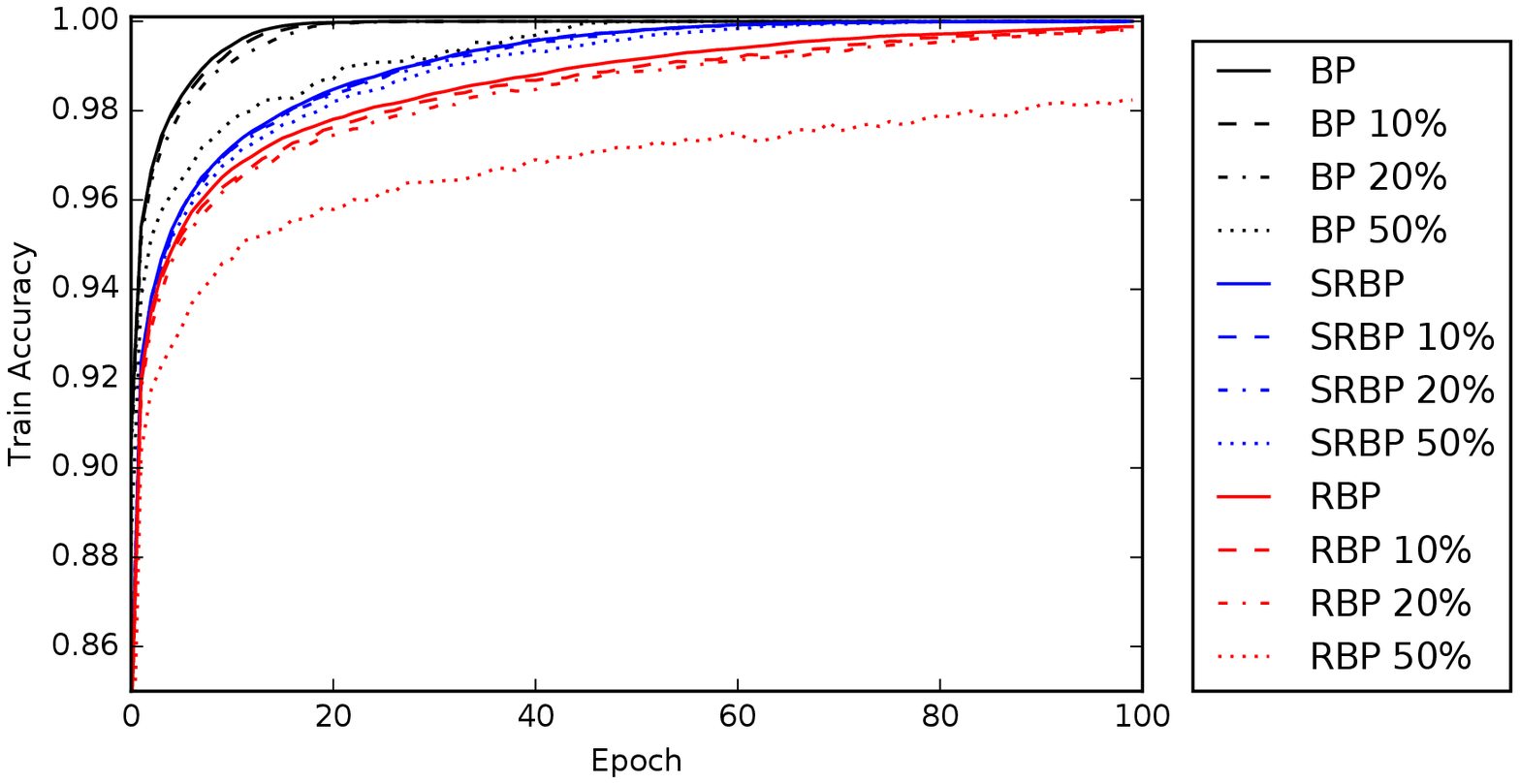}
\end{subfigure}
\begin{subfigure}[b]{\textwidth}
\includegraphics[width=\textwidth]{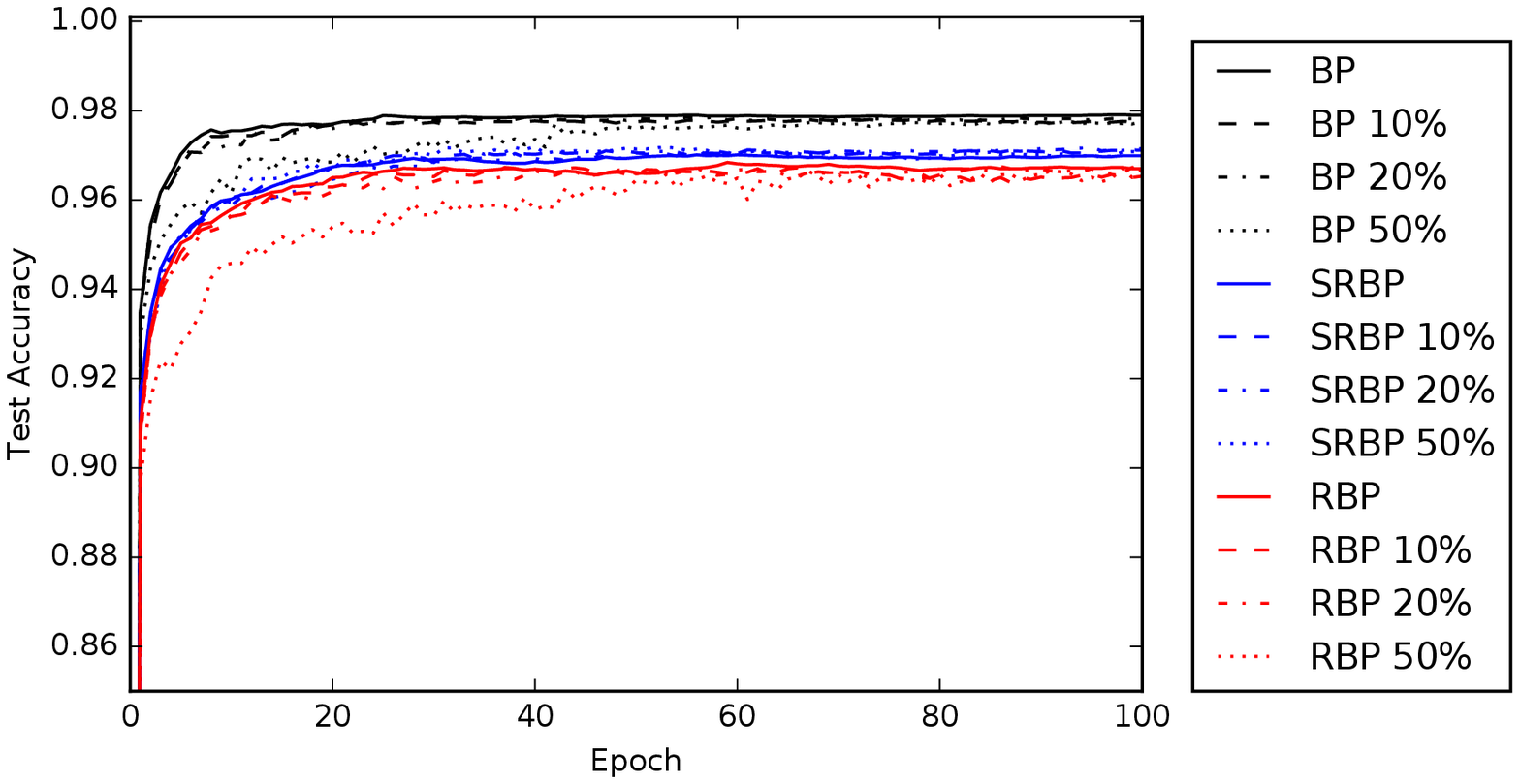}
\end{subfigure}
\caption{MNIST training (upper) and test (lower) accuracy, as a function of training epoch, for BP, RBP, and SRBP with different dropout probabilities in the learning channel: 0\% (no dropout), 10\%, 20\%, and 50\%. For dropout probability $p$, the error signals that are not dropped out are scaled by $1/(1-p)$. As with dropout in the forward propagation, large dropout probabilities lead to slower training without hurting final performance.}
\label{fig:dropout}
\end{figure}

Finally, it is possible to use dropout in the backward pass.
The forward pass is robust to dropping out neurons, and in fact the dropout procedure can be beneficial
\cite{hinton_improving_2012,baldidropout14}. Here we apply the dropout procedure to neurons in the learning channel during the backward pass. The results of simulations are reported
in Figure \ref{fig:dropout} and confirm that BP, RBP, SRBP, are robust with respect to dropout.

\section{Mathematical Analysis}

\subsection{General Considerations}

The general strategy to try to derive more precise mathematical results is to proceed from simple architectures to more complex architectures, and from the linear case to the non-linear case. The linear case is more amenable to analysis and, in this case, RBP and SRBP are equivalent when
there is only one hidden layer, or when all the layers have the same size. Thus we study the convergence of RBP to optimal solutions in linear architectures of increasing complexity: ${\cal A}[1,1,1]$,  ${\cal A}[1,1,1,1]$,
${\cal A}[1,1,\ldots,1]$, ${\cal A}[1,N,1]$
 ${\cal A}[N,1,N]$, and then the general  ${\cal A}[N0,N1,N2]$ case with a single hidden layer. This is followed by the study of a non-linear ${\cal A}[1,1,1]$ case.

For each kind of linear network, under a set of standard assumptions, one ca derive a set of non-linear--in fact polynomial--autonomous, ordinary differential equations (ODEs) for the average (or batch) time evolution of the synaptic weights under the RBP or SRBP algorithm. 
As soon as there is more than one variable and the system is non-linear, there is no general theory to understand the corresponding behavior. In fact, even in two dimensions,
the problem of understanding the upper bound on the number and relative position of the limit cycles of a system of the form $dx/dt=P(x,y)$ and $dy/dt=Q(x,y)$, where $P$ and $Q$ are polynomials of degree $n$ is open--in fact this is Hilbert's 16-th problem in the field of dynamical systems
\cite{smale1998mathematical,ilyashenko2002centennial}.

When considering the specific systems arising from the RBP/SRBP learning equations, one must first prove that these systems have a long-term solution. Note that polynomial ODEs may not have long-term solutions (e.g. $dx/dt=x^\alpha$, with $x(0)\not = 0$, does not have long-term solutions for $\alpha>1$). If the trajectories are bounded, then long-term solutions exist. We are particularly interested in long-term solutions that converge to a fixed point, as opposed to limit cycles or other behaviors. 

A number of interesting cases can be reduced to polynomial differential equations in one dimension. These can be understood using the following theorem.

\par\null
\noindent
{\bf Theorem 1:}
Let $dx/dt=Q(x)=k_0+k_1x+\ldots +k_nx^n$ be a first order polynomial differential equation in one dimension of degree $n>1$, and let 
$r_1<r_2 \ldots <r_k$ ($k \leq n$) be the ordered list of distinct real roots of $Q$ (the fixed points). If $x(0)=r_i$ then $x(t) =r_i$ and the solution is constant  If $r_i<x(0)<r_{i+1}$ then $x(t) \to r_i$ if $Q<0$ in $(r_i,r_{i+1})$, and 
$x(t) \to r_{i+1}$ if $Q>0$ in $(r_i,r_{i+1})$. If $x(0)<r_1$ and $Q>0$ in the corresponding interval, then 
$x(t) \to r_1$. Otherwise, if $Q<0$ in the corresponding interval, there is no long time solution and $x(t)$ diverges to $- \infty$ within a finite horizon.
If $x(0)>r_k$ and $Q<0$ in the corresponding interval, then 
$x(t) \to r_k$. Otherwise, if $Q>0$ in the corresponding interval, there is no long time solution and $x(t)$ diverges to $+ \infty$ within a finite horizon.
{\it A necessary and sufficient condition for the dynamics to always converge to a fixed point is that the degree $n$ be odd, and the leading coefficient be negative.}

\par \null
\noindent
{\bf Proof:} The proof of this theorem is easy and can be visualized by plotting the function $Q$.

Finally, in general the matrices in the forward channel are denoted by $A_1, A_2, \dots$, and the matrices in the learning channel are denoted by $C_1, C_2, \ldots$
Theorems are stated in concise form and additional important facts are contained in the proofs.

\subsection{The Simplest Linear Chain:  ${\cal A}[1,1,1]$}

\begin{figure}[h!]
    \centering
    \includegraphics[width=0.95\textwidth]{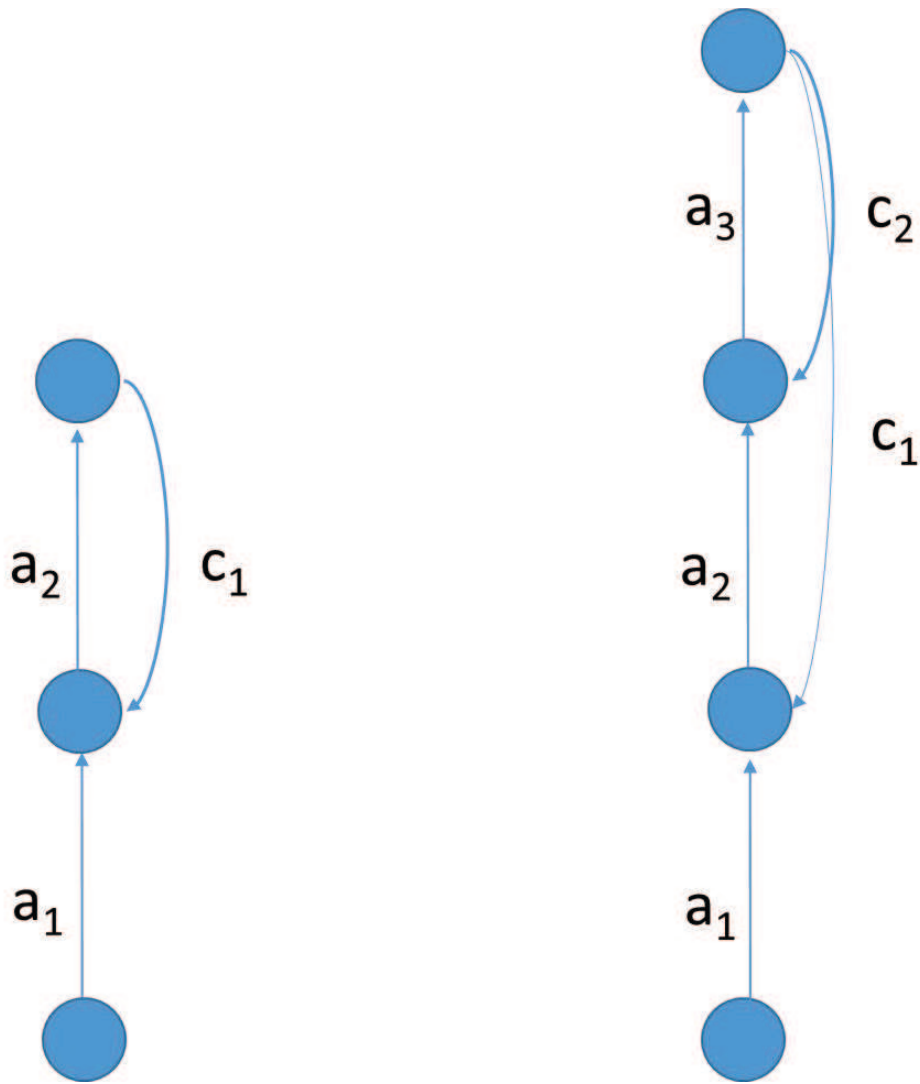}
    \caption{Left: ${\cal A}[1,1,1]$ architecture. The weights $a_1$ and $a_2$ are adjustable, and the feedback weight $c_1$ is constant. Right:  ${\cal A}[1,1,1,1]$ architecture. The weights $a_1,a_2$, and $a_3$ are adjustable, and the feedback weights $c_1$ and $c_2$ are constant.   
    }
    \label{fig:A111}
\end{figure}

{\bf Derivation of the System:} The simplest case correspond to a linear ${\cal A}[1,1,1]$ architecture (Figure \ref{fig:A111}). Let us denote by $a_1$ and $a_2$ the weights in the first and second layer, and by $c_1$ the random weight of the learning channel. In this case, we have $O(t)=a_1a_2I(t)$ and the learning equations are given by:

\be
\begin{cases}
\Delta a_1 = \eta c_1(T-O)I=\eta c_1(T-a_1a_2I)I\\
\Delta a_2 = \eta(T-O) a_1I = \eta (T-a_1a_2I) a_1I
\end{cases}
\label{eq:lin1}
\ee
When averaged over the training set:
\be
\begin{cases}
E(\Delta a_1 )= \eta c_1 E(IT) -\eta c_1a_1a_2 E(I^2)=\eta c_1 \alpha -\eta c_1a_1a_2 \beta\\
E(\Delta a_2 )= \eta a_1 E(IT)-\eta a_1^2a_2 E(I^2)=
\eta a_1 \alpha -\eta a_1^2a_2 \beta
\end{cases}
\label{eq:lin2}
\ee
where $\alpha=E(IT)$ and $\beta = E(I^2)$.
With the proper scaling of the learning rate ($\eta=\Delta t$) this leads to the non-linear system of coupled differential equations for the temporal evolution of $a_1$ and $a_2$ during learning:

\be
\begin{cases}
\frac{da_1}{dt}= \alpha c_1  - \beta c_1a_1a_2 =c_1(\alpha - \beta a_1a_2)\\
\frac{da_2}{dt} = \alpha a_1- \beta a_1^2a_2 = a_1(\alpha-\beta a_1a_2)
\end{cases}
\label{eq:lin3}
\ee
Note that the dynamic of $P=a_1a_2$ is given by:

\be
\frac{dP}{dt}=a_1\frac{da_2}{dt}+a_2\frac{da_1}{dt}=(a_1^2+a_2c_1)(\alpha-\beta P)
\label{eq:lin4}
\ee
The error is given by:

\be 
{\cal E}=\frac{1}{2} E(T-PI)^2=\frac{1}{2}E(T^2) + \frac{1}{2}P^2\beta -P \alpha=\frac{1}{2}E(T^2) +\frac{1}{2\beta}(\alpha-\beta P)^2-\frac{\alpha^2}{2\beta}
\label{eq:error1}
\ee
and:

\be 
\frac{d {\cal E}}{dP}=-\alpha+\beta P \quad
{\rm with} \quad \frac{\partial {\cal E}}{\partial a_i}=( -\alpha + \beta P )\frac{P}{a_i}
\label{eq:error2}
\ee
the last equality requires $a_i \not = 0$.

\null\par
\noindent
{\bf Theorem 2:} The system in Equation \ref{eq:lin3} always converges to a fixed point. Furthermore, except for trivial cases associated with $c_1=0$, starting from any initial conditions the system converges to a fixed point corresponding to a global minimum of the quadratic error function. All the fixed points are located on the hyperbolas given by $\alpha-\beta P=0$ and are global minima of the error function. All the fixed points are attractors except those that are interior to a certain parabola. For any starting point, the final fixed point can be calculated by solving a cubic equation.

\null\par
\noindent{\bf Proof:}
As this is the first example, we first deal with the trivial cases in detail. For subsequent systems, we will skip the trivial cases entirely.

\noindent
{\it Trivial Cases:}
1) If $\beta=0$ then we must have $I=0$ and thus $\alpha=0$. As a result the activity of the input, hidden, and output, neuron will always be 0. Therefore the weights $a_1$ and $a_2$ will remain constant ($da_1/dt=da_2/dt=0$) and equal to their initial values $a_1(0)$ and $a_2(0)$. The error will also remain constant, and equal to 0 if and only if $T=0$.
Thus from now on we can assume that 
$\beta > 0$.

\noindent
2) If $c_1=0$ then the lower weight $a_1$ never changes and remains equal to its initial value. If this initial value
satisfies $a_1(0)=0$, then the activity of the hidden and output unit remains equal to 0 at all times, and thus $a_2$ remains constant and equal to its initial value $a_2=a_2(0)$. The error remains constant and equal to 0 if only if $T$ is always $0$.
If $a_1(0)\not = 0$, then the error is a simple quadratic convex function of $a_2$ and since the rule for adjusting $a_2$ is simply gradient descent, the value of $a_2$ will converge to its optimal value given by: $a_2=\alpha/\beta a_1(0)$. 

\noindent
{\it General Case:} Thus from now on, we can assume that $\beta > 0$ and $c_1 \not = 0$. 
Furthermore, it is easy to check that changing the sign of $\alpha$ corresponds to a reflection about the $a_2$-axis. Likewise, changing the sign of $c_1$ corresponds to 
a reflection about the origin (i.e. across both the $a_1$ and $a_2$ axis). Thus in short, it is sufficient to focus on the case where: $\alpha>0$, $\beta>0$, and $c_1>0$.

\begin{figure}[h!]
    \centering
   \includegraphics[width=1.0\textwidth]{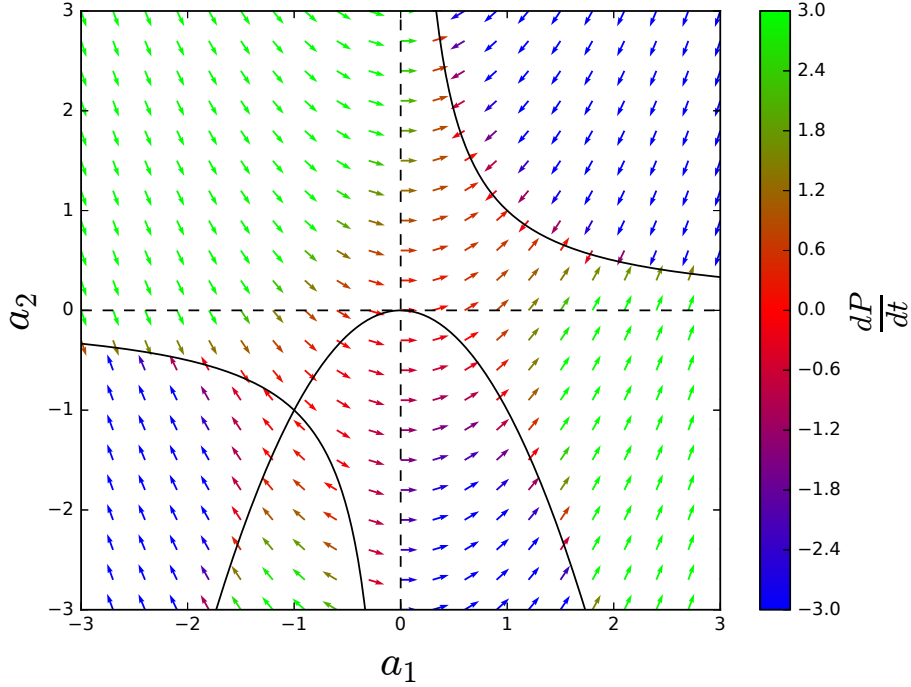}
    \caption{Vector field for the ${\cal A}[1,1,1]$ linear case with $c_1=1$, $\alpha=1$, and $\beta=1$. $a_1$ correspond to the horizontal axis and $a_2$ correspond to the vertical axis. The critical points correspond to the two hyperbolas, and all critical points are fixed points and global minima of the error functions. Arrows are colored according to the value of $dP/dt$, showing how the critical points inside the parabola $a_2=-a_1^2/c_1$ are unstable. All other critical points are attractors. Reversing the sign of $\alpha$, leads to a reflection across the $a_2$-axis; reversing the sign of $c_1$, leads to a reflection across both the $a_1$ and $a_2$ axes. }
    \label{fig:flow}
\end{figure}

\noindent 
In this case, the critical points for $a_1$ and $a_2$ are given by:

\be
P=a_1a_2=\frac{\alpha}{\beta} =\frac{E(IT)}{E(I^2)}=0
\label{eq:lin5}
\ee
which corresponds to two hyperbolas in the two-dimensional $(a_1,a_2)$ plane, in the first and third quadrant for $\alpha=E(IT)>0$.
{\it Note that these critical points do not depend on the feedback weight $c_1$}.
All these critical points correspond to global minima of the error function ${\cal E}={1 \over 2} E[(T-O)^2]$.
Furthermore, the critical points of $P$ include also the parabola:

\be a_1^2+a_2c_1=0 \quad {\rm or} \quad  a_2=-a_1^2/c_1
\label{eq:parabola100}
\ee
(Figure \ref{fig:flow}). 
These critical points are dependent on the weights in the learning channel. This parabola intersects the hyperbola $a_1a_2=P=\alpha/\beta$ at one point with coordinates: $a_1=(-c_1 \alpha/\beta)^{1/3}$   and $a_2=(-\alpha^{2/3}/(c_1^{1/3} \beta^{2/3})$.

In the upper half plane, where $a_2$ and $c_1$ are congruent and both positive, the dynamics is  simple to understand. For instance in the first quadrant where $a_1,a_2,c_1> 0$, if $\alpha-\beta P>0$ then $da_1/dt >0$,
$da_2/dt>0$, and $dP/dt>0$ everywhere and therefore the gradient vector flow is directed towards the hyperbola of critical points. If started in this region, $a_1$, $a_2$, and $P$ will grow monotonically until a critical point is reached and the error will decrease monotonically towards a global minimum.
If 
$\alpha-\beta P<0$ then $da_1/dt <0$,
$da_2/dt<0$, and $dP/dt<0$ everywhere and again the vector flow is directed towards the hyperbola of critical points. If started in this region, $a_1$, $a_2$, and $P$ will decrease monotonically until a critical point is reached and the error will decrease monotonically towards a global minimum. A similar situation is observed in the fourth quadrant where $a_1<0$ and $a_2>0$.

More generally, if $a_2$ and $c_1$ have the same sign, i.e. are congruent as in BP, then 
$a_1^2+a_2c_1 \geq 0$ and $P$ will increase if 
$\alpha -\beta P >0$, and decrease if $\alpha -\beta P <0$. Note however that this is also true in general when $c_1$ is small regardless of its sign, relative to $a_1$ and $a_2$,  since in this case it is still true that 
$a_1^2+a_2c_1$ is positive. This remains true even if $c_1$ varies, as long as it is small. When $c_1$ is small, the dynamics is dominated by the top layer. The lower layer changes slowly and the top layer adapts rapidly so that the system again converges to a global minimum.
When $a_2=c_1$ one recovers the convergent dynamic of BP, as $dP/dt$ always has the same sign as $\alpha-\beta P$.
However, in the lower half plane, the situation is slightly more complicated (Figure \ref{fig:flow}). 

To solve the dynamics in the general case, from Equation \ref{eq:lin3} we get:

\be
\frac{da_2}{dt}=a_1\frac{1}{c_1} \frac {da_1}{dt}
\label{eq:lin20}
\ee
which gives $a_2=\frac{1}{2c_1} a_1^2 +C$ so that finally:

\be
a_2=\frac{1}{2c_1} a_1^2 +b(0)-\frac{1}{2c_1}a_1^2(0) 
\label{eq:lin21}
\ee
Given a starting point $a_1(0)$ and $a_2(0)$, the system will follow a trajectory given by the parabola in Equation \ref{eq:lin21} until it converges to a critical point (global optimum) where $da_1/dt=da_2/dt=0$. To find the specific critical point to which it converges to, Equations \ref{eq:lin21}
and \ref{eq:lin5} must be satisfied simultaneously which leads to the depressed cubic equation:

\be
a_1^3+(2c_1a_2(0)-a_1(0)^2)a_1-2\frac{c_1\alpha}{\beta}=0
\label{eq:lin22}
\ee
which can be solved using the standard formula for the roots of cubic equations. 
Note that the parabolic trajectories contained in the upper half plane intersect the critical hyperbola in only one point and therefore the equation has a single real root. In the lower half plane, the parabolas associated with the trajectories can intersect the hyperbolas in 1, 2, or 3 distinct points corresponding to 1 real root, 2 real roots (1 being double), and 3 real roots. The double root corresponds to the point
$-(c_1 \alpha/\beta)^{1/3}$ associated with the intersection of the parabola of Equation \ref{eq:lin21} with both the hyperbola of critical points $a_1a_2=\alpha/\beta$ and the parabola of additional critical points for $P$ given by Equation \ref{eq:parabola100}.

When there are multiple roots, the convergence point of each trajectory is easily identified by looking at the derivative vector flow (Figure \ref{fig:flow}). Note on the figure that all the points on the critical hyperbolas are stable attractors, except for those in the lower half-plane that satisfy both $a_1a_2=\alpha/\beta$ and $a_2c_1+a_1^2<0$. 
This can be shown by linearizing the system around its critical points. 

\noindent
{\bf Linearization Around Critical Points.}
If we consider a small deviation $a_1+u$ and $a_2+v$ around a critical point $a_1,a_2$ (satisfying $\alpha -\beta a_1a_2=0$)
and linearize the corresponding system, we get:

\be
\begin{cases}
\frac{du}{dt}= -\beta c_1(a_2u+a_1v) \\
\frac{dv}{dt} = -\beta a_1( a_2u + a_1v)
\end{cases}
\label{eq:lin6}
\ee
with $a_1a_2=\alpha/\beta$. If we let $w=a_2u+a_1v$ we have:

\be
\frac{dw}{dt}=-\beta(c_1a_2+a_1^2)w \quad {\rm thus} \quad
w=w(0)e^{-\beta(c_1a_2+a_1^2) t}
\label {eq:lin7}
\ee
Thus if $\beta(c_1a_2+a_1^2)>0$, $w$ converges to zero and $a_1,a_2$ is an attractor. In particular, this is always the case when $c_1$ is very small, or $c_1$ has the same sign as $a_2$.
If $\beta(c_1a_2+a_1^2)<0$, $w$ diverges to $+\infty$, and corresponds to unstable critical points as described above.
If $\beta(c_1a_2+a_1^2)=0$, $w$ is constant. 

Finally, note that in many cases, for instance for trajectories in the upper half plane, the value of $P$ along the trajectories increases or decreases monotonically towards the global optimum value. However this is not always the case and there are trajectories where $dP/dt$ changes sign, but this can happen only once.

\subsection{Adding Depth: the Linear Chain ${\cal A}[1,1,1,1]$.}

{\bf Derivation of the System:} In the case of a linear ${\cal A}[1,1,1,1]$ architecture, for notational simplicity, let us denote by $a_1, a_2$ and $a_3$ the forward weights, and by $c_1$ and $c_2$ the random weights of the learning channel (note the index is equal to the target layer). In this case, we have $O(t)=a_1a_2a_3I(t)=PI(t)$. The learning equations are:

\be
\begin{cases}
\Delta a_1 = \eta c_1(T-O)I=\eta c_1(T-a_1a_2a_3I)I\\
\Delta a_2 = \eta c_2(T-O) a_1I = \eta c_2 (T-a_1a_2a_3I) a_1I\\
\Delta a_3= \eta (T-O)a_1a_2I=\eta (T-a_1a_2a_3I)a_1a_2I
\end{cases}
\label{eq:linc1}
\ee
When averaged over the training set:

\be
\begin{cases}
E(\Delta a_1 )= \eta c_1  E(IT) -\eta c_1P E(I^2)=\eta c_1 \alpha -\eta c_1 P \beta\\
E(\Delta a_2 )= \eta c_2 a_1  E(IT)-\eta c_2a_1 P E(I^2)=
\eta c_2a_1 \alpha -\eta c_2a_1P \beta\\
E(\Delta a_3)=\eta a_1a_2 E(IT)-\eta a_1a_2PE(I^2)=\eta a_1a_2 \beta -\eta a_1a_2P \beta
\end{cases}
\label{eq:linc2}
\ee
where $P=a_1a_2a_3$.
With the proper scaling of the learning rate ($\eta=\Delta t$) this leads to the non-linear system of coupled differential equations for the temporal evolution of $a_1,a_2$ and $a_3$ during learning:

\be
\begin{cases}
\frac{da_1}{dt}= c_1(\alpha - \beta P)\\
\frac{da_2}{dt} = c_2a_1(\alpha-\beta P)\\
\frac{da_3}{dt}= a_1a_2(\alpha -\beta P)
\end{cases}
\label{eq:linc3}
\ee
The dynamic of $P=a_1a_2a_3$ is given by:

\be
\frac{dP}{dt}=a_1a_2\frac{da_3}{dt}+a_2a_3\frac{da_1}{dt}+a_1a_3 \frac{da_2}{dt}=
(a_1^2a_2^2 + c_1a_2a_3+ c_2 a_1^2a_3)(\alpha-\beta P)
\label{eq:linc4}
\ee

\par\null
\noindent
{\bf Theorem 3:} Except for trivial cases (associated with $c_1=0$ or $c_2=0$), starting from any initial conditions the system in Equation \ref{eq:linc3} converges to a fixed point, corresponding to a global minimum of the quadratic error function. All the fixed points are located on the hypersurface given by $\alpha-\beta P=0$ and are global minima of the error function. Along any trajectory, and for each $i$, $a_{i+1}$ is a quadratic function of $a_i$. For any starting point, the final fixed point can be calculated by solving a polynomial equation of degree seven.

\par\null
\noindent
{\bf Proof:}
If $c_1=0$, $a_1$ remains constant and thus we are back to the linear case of a ${\cal A}[1,1,1]$ architecture where the inputs $I$ are replaced by $a_1I$. Likewise, if $c_2=0$ $a_2$ remains constant and the problem can again be reduced to the ${\cal A}[1,1]$ case with the proper adjustments.
Thus for the rest of this section  we can assume $c_1 \neq 0$ and $c_2 \neq 0$.

The critical points of the system correspond to $\alpha - \beta P= 0$ and do not depend on the weights in the learning channel. These critical points correspond to global minima of the error function. These critical points are also critical points for the product $P$. Additional critical points for $P$ are provided by the hypersurface: 
$a_1^2a_2^2 + c_1a_2a_3+ c_2 a_1^2a_3=0$ with $(a_1,a_2,a_3)$ in $\mathbb R^3$.

The dynamics of the system can be solved by noting that Equation \ref{eq:linc3} yields:

\be
\frac{da_2}{dt}=\frac{a_1c_2}{c_1} \frac {da_1}{dt} \quad {\rm and} \quad \frac{da_3}{dt}=\frac{a_2}{c_2}  \frac{da_2}{dt}
\label{eq:linc5}
\ee
As a result: 

\be
a_2=\frac{c_2}{2c_1}a_1^2+C_1 \quad {\rm with} \quad 
C_1=  a_2(0)- \frac{c_2}{2c_1}a_1(0)^2
\label{eq:linc6}
\ee
and:

\be
a_3=\frac{1}{2c_2}a_2^2+C_2 \quad {\rm with} \quad
C_2=  a_3(0)- \frac{1}{2c_2}a_2(0)^2
\label{eq:linc7}
\ee
Substituting these results in the first equation of the system gives:

\be
\frac{da_1}{dt}=c_1[\alpha - \beta a_1(\frac{c_2}{2c_1}a_1^2+C_1)(   \frac{1}{2c_2}a_2^2+C_2                    )]
\label{eq:linc8}
\ee
and hence:

\be
\frac{da_1}{dt}=c_1[\alpha - \beta a_1(\frac{c_2}{2c_1}a_1^2+C_1)(   \frac{1}{2c_2}(\frac{c_2}{2c_1}a_1^2+C_1     )^2+C_2   )]
\label{eq:linc9}
\ee
In short $da_1/dt=Q(a_1)$ where $Q$ is a polynomial of degree 7 in $a_1$. By expanding and simplifying Equation 
\ref{eq:linc9}, it is easy to see that the leading term of $Q$ is negative and given by $\beta c_2^2/(16c_1^2)$. Therefore, using Theorem 1, for any initial conditions $a_1(0)$, $ a_1(t)$ converges to a finite fixed point. Since $a_2$ is a quadratic function of $a_1$ it also converges to a finite fixed point, and similarly for $a_3$.
Thus in the general case the system always converges to a global minimum of the error function satisfying $\alpha - \beta P=0$. 
The hypersurface  $a_1^2a_2^2 + c_1a_2a_3+ c_2 a_1^2a_3=0$ depends on $c_1,c_2$ and provides additional critical points for the product $P$. It can be shown again by linearization that this hypersurface separates stable from unstable fixed points.

As in the previous case, small weights and congruent weights can help learning but are not necessary. In particular, if  $c_1$ and $c_2$ are small, or if $c_1$ is small and $c_2$ is congruent (with $a_3$), then $a_1^2a_2^2 + c_1a_2a_3+ c_2 a_1^2a_3>0$ and $dP/dt$ has the same sign as 
$\alpha-\beta P$.

\subsection{The General Linear Chain: ${\cal A}[1, \dots,1]$.}

\begin{figure}[h!]
    \centering
   \includegraphics[width=0.40\textwidth]{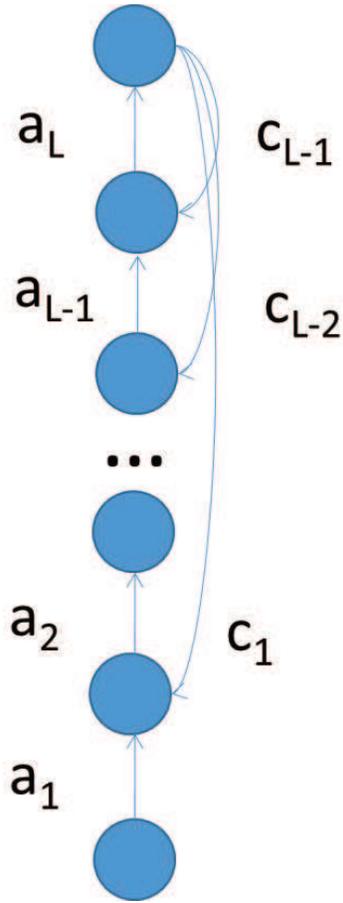}
    \caption{Left: ${\cal A}[1,\ldots,1]$ architecture. The weights $a_i$ are adjustable, and the feedback weight $c_i$ are fixed. The index of each parameter is associated with the corresponding target layer.   
    }
    \label{fig:A1111}
\end{figure}

{\bf Derivation of the System:}
The analysis can be extended immediately to a linear chain architecture ${\cal A}[1, \dots,1]$ of arbitrary length
(Figure \ref{fig:A1111}). In this case, let $a_1,a_2, \ldots, a_L$ denote the forward weights and $c_1, \dots ,c_{L-1}$ denote the feedback weights. 
Using the same derivation as in the previous cases and letting $O=PI=a_1 a_2 \ldots a_L I$ gives the system: 


\be
\Delta a_i=\eta c_i (T-O) a_1a_2 \ldots a_{i-1}I
\label{eq:}
\ee
for $i=1,\ldots ,L$.
Taking expectations as usual leads to the set of differential equations:
\be
\begin{cases}
\frac{da_1}{dt}= c_1(\alpha - \beta P)\\
\frac{da_2}{dt} =c_2a_1 (\alpha-\beta P)\\
\ldots \\
\frac{da_{L-1}}{dt}=c_{L-1}a_1a_2 \ldots a_{L-2} (\alpha -\beta P)\\
\frac{da_L}{dt}= a_1\ldots a_{L-1}(\alpha -\beta P)
\end{cases}
\label{eq:lind3}
\ee
or, in more compact form:

\be
\frac{da_i}{dt}=c_{i} \prod_{k=1}^{k=i-1} a_k  (\alpha -\beta P) \quad {\rm for} \quad i=1, \dots, L\\
\label{eq:lind4}
\ee
with $c_L=1$.
As usual, $P=\prod_{i=1}^L a_i$, $\alpha= E(TI)$,
and $\beta=E(I^2)$. 
A simple calculation yields:

\be
\frac{dP}{dt}=\sum_{i=1}^L\frac{P}{a_i}\frac{da_i}{dt}=
(\alpha -\beta P)\sum_{i=1}^L P\frac{c_i}{a_i} \prod_{k=1}^{i-1} a_k
\label{eq:lind5}
\ee
the last equality requiring $a_i \not = 0$ for every $i$.

\par\null
\noindent
{\bf Theorem 4:} Except for trivial cases, starting from any initial conditions the system in Equation \ref{eq:lind3} converges to a fixed point, corresponding to a global minimum of the quadratic error function. All the fixed points are located on the hypersurface given by $\alpha-\beta P=0$ and are global minima of the error function. Along any trajectory, and for each $i$, $a_{i+1}$ is a quadratic function of $a_i$. For any starting point, the final fixed point can be calculated by solving a polynomial equation of degree $2^L-1$.

\par\null
\noindent
{\bf Proof:}
Again, when all the weights in the learning channel are non zero, the critical points correspond to the curve $\alpha-\beta P=0$. These critical points are independent of the weights in the learning channel and correspond to global minima of the error function.
Additional critical points for the product $P=a_1 \ldots a_L$ are given by the surface 
$\sum_{i=1}^L P\frac{c_i}{a_i} \prod_{k=1}^{i-1} a_k   =0$.
These critical points are dependent on the weights in the learning channel. If the $c_i$ are small or congruent with the respective feedforward weights, then 
$\sum_{k=1}^{L} [\prod_{i \not = k} a_i   ][   c_{L-k} \prod_{j=1}^{j=k-1} a_j     ]   >0$ and 
$dP/dt$ has the same sign as 
$\alpha-\beta P$. Thus small or congruent weights can help the learning but they are not necessary.

To see the convergence, from Equation \ref{eq:lind4}, we have:

\be
c_i \frac{da_{i+1}}{dt}=c_{i+1}a_i \frac{da_i}{dt}
\label{eq:lind6}
\ee
Note that if one the derivatives $da_i/dt$ is zero, then they are all zero and thus there cannot be any limit cycles. Since in the general case all the $c_i$ are non zero, we have:

\be
a_{i+1}=\frac{c_{i+1}}{2c_i} a_i^2 +C
\label{eq:lind7}
\ee
showing that there is a quadratic relationship between 
$a_{i+1}$ and $a_i$, with no linear term, for every $i$. Thus every $a_i$ can be expressed as a polynomial function of $a_1$ of degree $2^{i-1}$, containing only even terms:

\be
a_i=k_0+k_1a_1^2+ \ldots + k_{i-1}a_1^{2^{i-1}}
\label{eq:lind8}
\ee
and:

\be
k_{i-1}= \frac{c_{i}}{2c_{i-1}}(\frac{c_{i-1}}{2c_{i-2}})^2(\frac{c_{i-2}}{2c_{i-3}})^4 \ldots (\frac{c_3}{2c_{2}})^{2^{i-1}}
\label{eq:lind9}
\ee
By substituting these relationships in the equation for the derivative of $a_1$, we get $da_1/dt=Q(a_1)$ where $Q$ is a polynomial with an odd degree $n$ given by:

\be
n=1+2+4+\ldots +2^{L-1}=2^L-1
\label{eq:}
\ee
Furthermore, from Equation \ref{eq:lind9} it can be seen that leading coefficient is negative therefore, using Theorem 1, for any set of initial conditions the system must converge to a finite fixed point. For a given initial condition, the point of convergence can be solved by looking at the nearby roots of the polynomial $Q$ of degree $n$.

\noindent
{\bf Gradient Descent Equations:}
For comparison, the gradient descent equations are:

\be
\frac{da_i}{dt}=a_L\ldots a_{i+1} a_1 \ldots a_{i-1} (\alpha-\beta P)=\frac{P}{a_i} (\alpha-\beta P)=
-\frac{\partial{\cal E}}{\partial a_i}
\label{eq:}
\ee
(the equality in the middle requires that $a_i \not = 0$). 
In this case, the coupling between neighboring terms is given by:

\be
a_i \frac{da_i}{dt}=a_{i+1} \frac{da_{i+1}}{dt}
\label{eq:}
\ee
Solving this equation yields:

\be
\frac{da_i^2}{dt}=\frac{da_{i+1}^2}{dt}
\quad {\rm or} \quad a_{i+1}^2=a_i^2 +C
\label{eq:}
\ee

\subsection{Adding Width (Expansive): ${\cal A}[1,N,1]$ }

\begin{figure}[h!]
    \centering
    \includegraphics[width=0.95\textwidth]{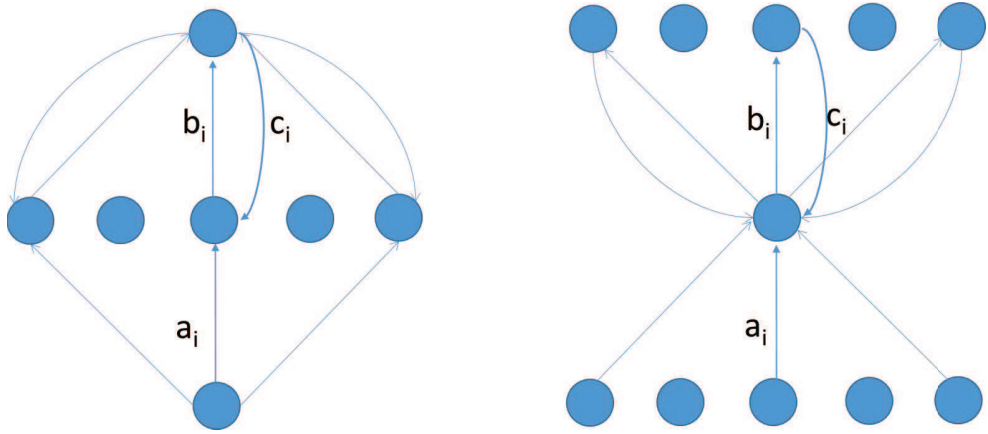}
    \caption{Left: Expansive ${\cal A}[1,N,1]$ Architecture. Right: Compressive ${\cal A}[N,1N]$ Architecture. In both cases, the parameters $a_i$ and $b_i$ are adjustable, and the parameters $c_i$ are fixed. }
    \label{fig:CompExp}
\end{figure}

{\bf Derivation of the System:}
Consider a linear ${\cal A}[1,N,1]$ architecture (Figure \ref{fig:CompExp}). For notational simplicity, we let $a_1,\ldots ,a_N$ be the weights in the lower layer, $b_1,\ldots ,b_N$ be the weights in the upper layer, and $c_1,\ldots ,c_N$ the random weights of the learning channel.
In this case, we have $O(t)=\sum_i a_ib_iI(t)$. We let $P=\sum_i a_ib_i$. The learning equations are:

\be
\begin{cases}
\Delta a_i = \eta c_i(T-O)I=\eta c_i(T-\sum_i a_ib_iI)I\\
\Delta b_i = \eta(T-O) a_iI = \eta (T-\sum_i a_ib_iI) a_iI
\end{cases}
\label{eq:linb1}
\ee
When averaged over the training set:
\be
\begin{cases}
E(\Delta a_i )= \eta c_i E(IT) -\eta c_i P E(I^2)=\eta c_i \alpha -\eta c_iP \beta\\
E(\Delta b_i )= \eta a_i E(IT)-\eta a_i P  E(I^2)=
\eta a_i \alpha -\eta a_iP \beta
\end{cases}
\label{eq:linb2}
\ee
where $\alpha=E(IT)$ and $\beta = E(I^2)$.
With the proper scaling of the learning rate ($\eta=\Delta t$) this leads to the non-linear system of coupled differential equations for the temporal evolution of $a_i$ and $b_i$ during learning:

\be
\begin{cases}
\frac{da_i}{dt}= \alpha c_i  - \beta c_iP =c_i(\alpha - \beta P)\\
\frac{db_i}{dt} = \alpha a_i- \beta a_iP = a_i(\alpha-\beta P)
\end{cases}
\label{eq:linb3}
\ee
The dynamic of $P=\sum_i a_ib_i$ is given by:

\be
\frac{dP}{dt}=\sum_i a_i\frac{db_i}{dt}+b_i\frac{da_i}{dt}=(\alpha-\beta P)\sum_i[b_ic_i+a_i^2]
\label{eq:linb4}
\ee

\par\null
\noindent
{\bf Theorem 5:} Except for trivial cases, starting from any initial conditions the system in Equation \ref{eq:linb3} converges to a fixed point, corresponding to a global minimum of the quadratic error function. All the fixed points are located on the hyersurface given by $\alpha-\beta P=0$ and are global minima of the error function. Along any trajectory, each $b_i$ is a quadratic polynomial function of $a_i$. Each $a_i$ is an affine function of any other $a_j$.For any starting point, the final fixed point can be calculated by solving a polynomial differential equation of degree $3$. 

\par\null
\noindent
{\bf Proof:}
Many of the features found in the linear chain are found again in this system using similar analyses.
In the general case where the weights in the learning channel are non zero, the critical points are given by the surface 
$\alpha-\beta P=0$ and correspond to global optima.
These critical points are independent of the weights in the learning channel.
Additional critical points for the product $P=\sum_i a_ib_i$ are given by the surface
$\sum_i a_i^2 + b_ic_i =0$ which depends on the weights in the learning channel.
If the $c_i$'s are small, or congruent with the respective $b_i$'s, then 
$\sum_i a_i^2 + b_ic_i >0$ and $dP/dt$ has the same sign as $\alpha-\beta P$.

To address the convergence, Equations \ref{eq:linb3} leads to the vertical coupling between $a_i$ and $b_i$:
\be
a_i \frac{da_i}{dt}=c_i \frac{db_i}{dt} \quad {\rm or} \quad
b_i=\frac{1}{2c_i}a_i^2 +C_i
\label{eq:h2}
\ee
for each $i=1,\ldots,N$.
Thus the dynamics of the $a_i$ variables completely determines the dynamics of the $b_i$ variables, and one only needs to understand the behavior of the $a_i$ variables.
In addition to the vertical coupling between $a_i$ and $b_i$, there is an horizontal coupling between the $a_i$ variables given again by Equation \ref{eq:linb3} resulting in:

\be
\frac{da_{i+1}}{dt}=\frac{c_{i+1}}{c_i}\frac{da_i}{dt} \quad
{\rm or} \quad a_{i+1}=\frac{c_{i+1}}{c_i}a_i+K_{i+1}
\label{eq:h4}
\ee
Thus, iterating, all the variables $a_i$ can be expressed as affine functions of $a_1$ in the form:

\be
a_i=\frac{c_i}{c_1}a_1+K'_i \quad i=1,\ldots,N
\label{eq:h5}
\ee
Thus solving the entire system can be reduced to solving for $a_1$. The differential equation for $a_1$ is of the form
$da_1/dt=Q(a_1)$ where $Q$ is a polynomial of degree 3.
Its leading term, is the leading term of $-c_1\beta P$.
To find its leading term we have:

\be 
P=\sum_ia_ib_i=\sum_i \frac{a_i^3}{2c_i}+c_ia_i
\label{eq:}
\ee
and thus the leading term of $Q$ is given by $Ka_1^3$ where:

\be 
K= -\beta c_1[\frac{1}{2c_1}+\frac{1}{2c_2}\frac{c_2^3}{c_1^3}+ \ldots \frac{1}{2c_N}\frac{c_N^3}{c_1^3}]=-\frac{\beta}{2}\frac{1}{c_1^2}[\sum_1^N c_i^2]
\label{eq:}
\ee
Thus the leading term of $Q$ has a negative coefficient, and therefore $a_1$ always converges to a finite fixed point, and so do all the other variables.

\subsection{Adding Width (Compressive): ${\cal A}[N,1,N]$ }

{\bf Derivation of the System:} 
Consider a linear ${\cal A}[N,1,N]$ architecture (Figure \ref{fig:CompExp}). The on-line learning equations are given by:

\be
\begin{cases}
\Delta a_i= \eta \sum_{k=1}^N c_k(T_k-O_k)I_i
\\
\Delta b_i= \eta (T_i-O_i) \sum_{k=1}^N a_kI_k
\end{cases}
\label{eq:}
\ee
for $i=1,\ldots,N$. As usual taking expectations, using matrix notation and a small learning rate, leads to the system of differential equations:

\be
\begin{cases}
\frac{dA}{dt}= C(\Sigma_{TI}-BA \Sigma_{II})
\\
\frac{dB}{dt}= (\Sigma_{TI}-BA \Sigma_{II})A^t
\end{cases}
\label{eq:333}
\ee
Here $A$ is an $1 \times N$ matrix, $B$ is an $N \times 1$ matrix, and $C$ is an $1 \times N$ matrix, and $M^t$ denotes the transpose of the matrix $M$. $\Sigma_{II}=E(II^t)$ and 
$\Sigma_{TI}=E(TI^t)$ are $N \times N$ matrices associated with the data.

\par \null
\noindent
{\bf Lemma 1:} Along the flow of the system defined by Equation \ref{eq:333}, the solution satisfies:

\be
CB={1 \over 2} \vert \vert A \vert \vert^2 + K
\label{eq:compressive0}
\ee
where $K$ is a constant depending only on the initial values.

\par\null
\noindent
{\bf Proof:} The proof is immediate since:

\be
C\frac{dB}{dt}=\frac{dA}{dt}A^t \quad {\rm or} \quad \sum_i c_i \frac{db_i}{dt}=\sum_i a_i \frac{da_i}{dt} \quad {\rm or}  \sum_i c_i \frac{db_i}{dt}=\frac{1}{2} \frac{d\vert\vert A \vert \vert^2}{dt}
\label{eq:compressive1}
\ee
where $\vert \vert A \vert \vert^2=a_1^2+\ldots +a_N^2$. 
The theorem is obtained by integration.

\par\null
\noindent
{\bf Theorem 6:} In the case of an autoencoder with uncorrelated normalized data (Equation \ref{eq:auto1000}), the system converges to a fixed point satisfying $A=\beta C$, where $\beta$ is a positive root of a particular cubic equation. At the fixed point,
 $B=C^t/(\beta \vert \vert C \vert\vert^2)$ and the product $P=BA$ converges to $C^tC/\vert\vert C\vert \vert^2$.

\par\null
\noindent
{\bf Proof:}
For an autoencoder
 with uncorrelated and normalized data ($\Sigma_{TI}=\Sigma _{II}=Id$). In this case the system can be written as:
 
 \be
 \begin{cases}
\frac{dA}{dt}=C(Id-BA)\\
\frac{dB}{dt}=(Id-BA)A^t
\label{eq:auto1000}
\end{cases}
\ee
We define 

\be
\sigma(t)=\frac 12\|A\|^2+ K
\label{eq:compressive3}
\ee
and let $A_0=A(0)$. Note that $\sigma(t) \geq K$.
We assume that $C$ and $A_0$ are linearly independent, otherwise the proof is easier.  Then we have:

\be
\frac{dA}{dt}=C-\sigma(t)A
\label{eq:compressive4}
\ee
Therefore the solution $A(t)$ must have the form:

\be
A(t)=f(t)C+g(t)A_0
\label{eq:compressive5}
\ee
which yields:

\be
\begin{split}
&f'(t)=1-\sigma(t)f(t),\quad f(0)=0
\\
&g'(t)=-\sigma(t)g(t), \quad g(0)=1
\end{split}
\label{eq:compressive6}
\ee

or:
\be
\begin{split}
& g(t)=e^{-\int_0^t\sigma(s)ds}\\
& f(t)=e^{-\int_0^t\sigma(s)ds}\int_0^t e^{\int_0^r\sigma(s)ds}dr
\end{split}
\label{eq:compressive7}
\ee
From the above expressions, we know that both $f$ and $g$ are nonnegative. We also have

\be
f(t)=g(t)\int_0^t\frac{1}{g(r)}dr
\label{eq:compressive8}
\ee
Since $\sigma (t) \geq K$, $g(t) $ is bounded, and thus 

\be
\int_0^\infty \frac{1}{g(r)}dr=\infty.
\label{eq:compressive9}
\ee
By a more general theorem shown in the next section, we know also that 
$\vert \vert A \vert \vert$ is bounded and therefore $f$ is also bounded. 
Using Equation \ref{eq:compressive8}, this implies that
that $g(t)\to 0$ as $t\to\infty$. Now we consider again the equation:

\be
f'=1-\sigma f
\label{eq:76}
\ee
Now consider the cubic equation:

\be
1-(\frac 12 t^2\vert \vert C \vert \vert^2 +K) t=0
\label{eq:77}
\ee
For $t$ large enough, since $g(t) \to 0$, we have:

\be
\sigma(t) \approx \frac{1}{2}f^2\vert \vert V \vert \vert^2 +K
\label{eq:}
\ee
Thus Equation \ref{eq:76} is close to the polynomial differential equation:

\be
h'=1-(\frac{1}{2}h^2 \vert \vert C \vert \vert^2+K)h
\label{eq:}
\ee
By Theorem 1, this system is always convergent to a positive root of Equation \ref{eq:77}, and by comparison the system in Equation \ref{eq:76} must converge as well.
This proves that $f(t)\to\beta$ as $t\to\infty$, and in combination with  $g(t)\to 0$ as  $t\to\infty$, shows that $A$ converges to $\beta C$. As $A$ converges to a fixed point, the error function converges to a convex function and $B$ performs gradient descent on this convex function and thus must also approach a fixed point. By the results in 
\cite{baldi88,baldicomplex12}, the solution must satisfy $BAA^t=A^t$. When $A=\beta C$ this gives: $B=C^t/(\beta CC^t)=C^t/(\beta \vert \vert C \vert\vert^2)$. In this case, the product $P=BA$ converges to the fixed point: 
$C^tC/\vert\vert C\vert \vert^2$.
The proof can easily be adapted to the slightly more general case where $\Sigma_{II}$ is a diagonal matrix.

\subsection{The General Linear Case: ${\cal A}[N_0,N_1, \ldots, N_L]$ }

\begin{figure}[h!]
    \centering
   \includegraphics[width=0.60\textwidth]{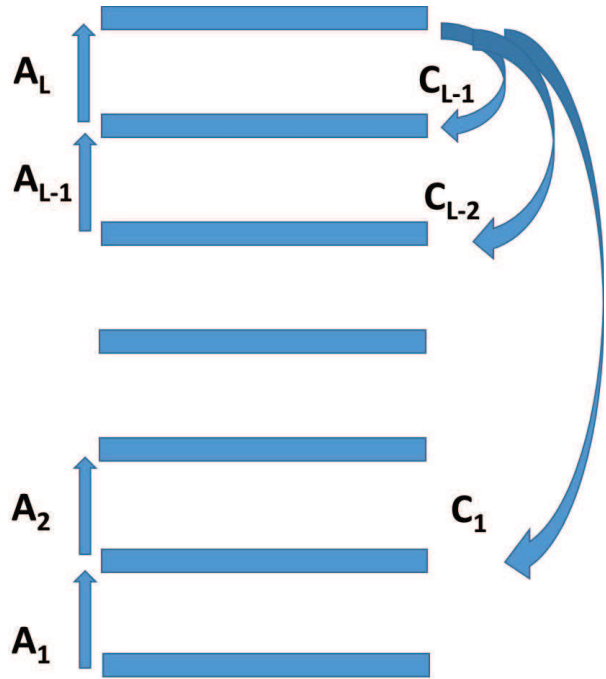}
    \caption{General linear case with an architecture 
    ${\cal A}[N_0, \ldots,N_L]$. Each forward matrix $A_i$ is adjustable and of size $N_i \times N_{i-1}$. In SRBP, each feedback matrices $C_i$ is fixed and of size $N_i \times N_L$.}
    \label{fig:GNL}
\end{figure}

{\bf Derivation of the System:}
Although we cannot yet provide a solution for this case, it is still useful to derive its equations. We assume a general feedforward linear architecture (Figure \ref {fig:GNL}) ${\cal A}[N_0,N_1,\ldots, N_L]$ with adjustable forward matrices $A_1, \ldots, A_L$ and fixed 
feedback matrices $C_1, \ldots,C_{L-1}$  (and $C_L=Id$).
Each matrix $A_i$ is of size $N_i \times N_{i-1}$ and, in SRBP, each matrix $C_i$ is of size $N_i \times N_L$.
As usual, $O(t)=PI(t)=(\prod_{i=1}^LA_i)I(t)$.

Assuming the same learning rate everywhere, using matrix notation we have:

\be
\Delta A_i =\eta C_i (T-O) (A_{i-1} \ldots A_1 I)t=\eta C_i(T-O)I^tA_1^t \ldots A_{i-1}^t
\label{eq:}
\ee
which, after taking averages, leads to the system of differential equations

\be
\frac{d A_i}{dt} = C_i (\Sigma_{TI} - P \Sigma_{II}) A_1^t \ldots A_{i-1}^t
\label{eq:GNL10}
\ee
with $P=A_L A_{L-1} \ldots A_1$, $\Sigma_{TI}=E(TI^t)$, and $\Sigma_{II}=E(II^t)$. $ \Sigma_{TI}$ is a $ N_L \times N_0$ matrix and 
$\Sigma_{II}$ is a $N_0 \times N_0 $ matrix. In the case of an autoencoder, $T=I$ and therefore $\Sigma_{TI}=\Sigma_{II}$.
Equation \ref{eq:GNL10} is true also for $i=1$ and $i=L$ with $C_L=Id$ where $Id$ is the identity matrix. 
These equations establish a coupling between the layers so that:  

\be
\frac{d A_{i+1}}{dt} = C_{i+1} (\Sigma_{TI} - P \Sigma_{II}) A_1^t \ldots A_{i}^t
\label{eq:}
\ee
When the layers have the same sizes, the coupling can be written as: 

\be
C_{i+1}^{-1}\frac{d A_{i+1}}{dt} = C_{i}^{-1} \frac{dA_i}{dt} A_{i}^t \quad {\rm or} \quad 
\frac{d A_{i+1}}{dt}=C_{i+1}C_{i}^{-1} \frac{dA_i}{dt} A_{i}^t 
\label{eq:}
\ee
where we can assume that the random matrices $C_i$ are invertible square matrices. 

\par\null
\noindent
{\bf Gradient Descent Equations:} For comparison,  the gradient descent equations are given by:

\be
\frac{dA_i}{dt}=A_{i+1}^t \ldots A_L^t
(\Sigma_{TI} - P \Sigma_{II}) A_1^t \ldots A_{i-1}^t
\label{eq:}
\ee
resulting in the coupling:

\be
A_{i+1}^t \frac {A_{i+1}}{dt}=\frac{dA_i}{dt}A_i^t 
\label{eq:}
\ee
and, by definition:

\be
\frac{dA_i}{dt}=-\frac{\partial {\cal E}}{\partial A_i}
\label{eq:}
\ee
where ${\cal E}=E(T-PI)^2/2$.

\par\null
\noindent
{\bf RBP Equations:} Note that in the case of RBP with backward matrices $C_1,
\ldots, C_{L-1}$, as opposed to SRBP, one has the system of differential equations:

\be
\frac{dA_i}{dt}=C_i \ldots C_{L-1}(\Sigma_{TI} - P \Sigma_{II})
A_1^t \ldots A_{i-1}^t
\label{eq:}
\ee
By letting $B_i=C_i \ldots C_{L-1}$ one obtains the SRBP equations however the size of the layers may impose contraints on the rank of the matrices $B_i$.

\subsection{ The General Three-Layer Linear Case ${\cal A}[N_0,N_1,N_2]$.}

{\bf Derivation of the System:} Here we let $A_1$ be the $N_1 \times N_0$ matrix of weights in the lower layer, $A_2$ be $N_2 \times N_1$ matrix of weights in the upper layer, and $C_1$ the $N_1 \times N_2$ random matrix of weights in the learning channel.
In this case, we have $O(t)=A_2A_1I(t)=PI(t))$ and $\Sigma_{II}=E(II^t) $ ($N_0 \times N_0 $), and $\Sigma_{TI}=E(TI^t) $ ($N_2 \times N_1 $). The learning equations are given by: 

\be
\begin{cases}
\frac{dA_2}{dt}= (\Sigma_{TI} - P \Sigma_{II})    A_1^t\\
\frac{dA_1}{dt}= C_1 (\Sigma_{TI} - P \Sigma_{II})   
\end{cases}
\label{eq:LGN3}
\ee
resulting in the coupling:
 
\be
C_1\frac{dA_2}{dt}= \frac{dA_1}{dt} A_1^t
\label{eq:GLN3S}
\ee
The corresponding gradient descent equations are obtained immediately by replacing $C_1$ with $A_2^t$.

Note that the two-layer linear case corresponds to the classical Least Square Method which is well understood.
The general theory of the three-layer linear case, however, is not well understood. In this section, we take a significant step towards providing a complete treatment of this case. One of the main results is that system defined by Equation \ref{eq:LGN3} has long-term existence, and $C_1P=C_1A_2A_1$ is convergent and thus, in short, the system is able to learn. However this alone does not imply that the matrix valued functions $A_1(t), A_2(t)$ are individually convergent. We can prove the latter in special cases like ${\cal A}[N,1,N]$ and ${\cal A}[1,N,1]$ studied in the previous sections, as well as ${\cal A}[2,2,2]$.

We begin with the following theorem.

\par\null
\noindent
{\bf Theorem 7:}
The general three layer linear system (Equation \ref{eq:LGN3}) always has long-term solutions. Moreover
$\vert \vert A_1 \vert \vert $ is  bounded.

\par\null
\noindent
{\bf Proof:} As in Lemma 1, we have:

\be
\frac{d(CA_2)}{dt}=C(\Sigma_{TI}-A_2A_1\Sigma_{II})A_1^t=\frac{dA_1}{dt}A_1^t
\label{eq:}
\ee
Thus we have:

\be
\frac{d((CA_2)+(CA_2)^t)}{dt}=\frac{dA_1}{dt}A_1^t+A_1\frac{dA_1^t}{dt}=\frac{d}{dt}(A_1A_1^t).
\label{eq:}
\ee
It follows that:

\be
(CA_2)+(CA_2)^t=A_1A_1^t+C_0
\label{eq:90}
\end{equation}
where $C_0$ is a constant matrix. Let:

\be
f={\rm Tr}(A_1A_1^t).
\label{eq:}
\ee
Using Lemma 2 below, we have:

\be
\frac{df}{dt}=2{\rm Tr}(\frac{dA_1}{dt}A_1^t)=2{\rm Tr}(C\Sigma_{TI}A_1^t-CA_2A_1\Sigma_{II} A_1^t)\leq c_3\|A_1\|-2{\rm Tr}(CA_2A_1\Sigma_{II} A_1^t).
\label{eq:}
\ee
Since:

\be
2{\rm Tr}(CA_2A_1\Sigma_{II}A_1^t)={\rm Tr}(CA_2A_1 \Sigma_{II}A_1^t) + {\rm Tr}(A_1\Sigma_{II}A_1^t(CA_2)^t)
\label{eq:}
\ee
or:

\be
2{\rm Tr}(CA_2A_1\Sigma_{II}A_1^t)={\rm Tr}(CA_2A_1 \Sigma_{II}A_1^t) + {\rm Tr}((CA_2)^tA_1\Sigma_{II}A_1^t)
\label{eq:}
\ee
using Equation \ref{eq:90}, we have:

\be
2{\rm Tr}(CA_2A_1\Sigma_{II}A_1^t)={\rm Tr}(A_1A_1^tA_1\Sigma_{II}A_1^t) +
{\rm Tr}(C_0A_1\Sigma_{II}A_1^t)
\label{eq:}
\ee
Using the second inequality in Lemma 2 below, we have:

\be
\frac{df}{dt} \leq c_3\vert\vert A_1\vert\vert+{\rm Tr}(C_0A_1\Sigma_{II}A_1^t)-c_1f^2 \leq c_3 \sqrt f + c_4 f -c_1f^2\leq c_5-\frac{1}{2}c_1f^2
\label{eq:}
\ee
for positive constants $c_1, \ldots, c_5$.
Since $A_1$ has long-term existence, so does $f$. Note that it is not possible for $f$ to be increasing as $t \to \infty$ because if we had $f'(t) \geq 0$, then we would have
$c_5-\frac{1}{2}c_1f^2 \geq0$ and thus $f$ would be bounded ($f \leq \sqrt{2c_5}/\sqrt{c_1}$). But if $f$ is not always increasing, at each local maximum point of $f$ we have
$f \leq \sqrt{2c_5}/\sqrt{c_1}$, which implies $ f \leq \sqrt{2c_5}/\sqrt{c_1}$
everywhere.

\par\null
\noindent
{\bf Lemma 2:}  There is  a constant $c_1>0$ such that
\begin{enumerate}
\item $f\geq c_1\|A_1\|^2$,
\item ${\rm Tr}(A_1A_1^tA_1\Sigma_{II} A_1^t)\geq c_1f^2$.
\end{enumerate}

\par\null
\noindent
{\bf Proof:}
The first statement is obvious. To prove the second one, we observe that:

\be
{\rm Tr}(A_1A_1^tA_1\Sigma_{II} A_1^t)={\rm Tr}( A_1^tA_1\Sigma_{II} A_1^tA_1)\geq c_2{\rm Tr}(A_1^tA_1A_1^tA_1)\geq c_1f^2
\label{eq:}
\ee
for some constants $c_1,c_2>0$.

\par\null
\noindent
To complete the proof of Theorem 7, we must estimate $A_2$ to make sure it does not diverge at a finite time. Let

\be
h=\frac{1}{2} {\rm Tr}(A_2A_2^t)
\label{eq:}
\ee
Then:

\be
\frac{dh}{dt}={\rm Tr}((\Sigma_{TI}-A_2A_1\Sigma_{II})A_1^tA_2^t)={\rm Tr}(\Sigma_{TI}A_1^tA_2^t)-
{\rm Tr}(A_2A_1\Sigma_{II}A_1^tA_2^t)
\label{eq:}
\ee
and thus:

\be
\frac{dh}{dt} \leq {\rm Tr}(\Sigma_{TI}A_1^tA_2^t)
\label{eq:}
\ee
Since we have shown that $\vert \vert A_1 \vert \vert$ is bounded:

\be
\frac{dh}{dt} \leq {\rm Tr}(\Sigma_{TI}A_1^tA_2^t) \leq K \vert \vert A_2 \vert\vert \leq K \sqrt h
\label{eq:}
\ee
for some constant $K$.
As a result, $h \leq K_1t^2 + K_2$ or

\be
\vert \vert A_2 \vert \vert \leq \sqrt{K_1t^2+K_2} \leq K_3t+K_4
\label{eq:}
\ee
Since for every $t$, $1/\vert\vert A_2 \vert \vert$ is bounded, the system has long-term solutions.

The main result of this section is as follows.
\par\null
\noindent
{\bf Theorem 8:}
[Partial Convergence Theorem] Along the flow of the system in Equation \ref{eq:LGN3}, $A_1$ and $C_1A_2$ are uniformly bounded. Moreover, $C_1A_2A_1\to C_1\Sigma_{TI}\Sigma_{II}^{-1}$ as $t\to\infty$ and:

\be
\int_0^\infty \|C_1A_2A_1- C_1\Sigma_{TI}\Sigma_{II}^{-1}\|^2dt<\infty
\label{eq:}
\ee

\par \null
\noindent
{\bf Proof.}
Let:

\be
U=C_1(\Sigma_{TI}-A_2A_1\Sigma_{II})\Sigma_{II}^{-1}
\label{eq:}
\ee
Then:

\be
\frac{dA_1}{dt}=U\Sigma_{II},\quad \frac{d(C_1A_2)}{dt} =U\Sigma_{II} A_1^T
\label{eq:}
\ee
It follows that:
\[
\begin{split}
&
\frac{d(C_1A_2)}{dt}(C_1A_2)^T=U\Sigma_{II} A_1^T(C_1A_2)^T=U\Sigma_{II}(C_1A_2A_1)^T\quad{\rm and}\\
&
\frac{d(A_1(C_1\Sigma_{TI}\Sigma_{II}^{-1})^T)}{dt}=U\Sigma_{II} (C_1\Sigma_{TI}\Sigma_{II}^{-1})^T=U(C_1\Sigma_{TI})^T
\end{split}
\]
Thus we have:

\begin{equation}\label{UU}
\frac{d(C_1A_2)}{dt}(C_1A_2)^T-\frac{d(A_1(C_1\Sigma_{TI}\Sigma_{II}^{-1})^T)}{dt}=-U\Sigma_{II} U^T\leq 0
\end{equation}
Here, for two matrices $X$ and $Y$, we write $X\leq Y$ if and only if $Y-X$ is a semi-positive matrix. 
Let:
\[
V=(C_1A_2)(C_1A_2)^T-A_1(C_1\Sigma_{TI}\Sigma_{II}^{-1})^T
-(C_1\Sigma_{TI}\Sigma_{II}^{-1})A_1^T
\]
Then: 
\[
\frac{dV}{dt}\leq 0
\]
By Theorem 7, there is a lower bound on the matrix $V$
\[
V\geq -A_1(C_1\Sigma_{TI}\Sigma_{II}^{-1})^T-
(C_1\Sigma_{TI}\Sigma_{II}^{-1})A_1^T
\geq -C
\]
for a constant matrix $C$. Thus as $t\to\infty$, $V=V(t)$ is convergent.
Using the inequality above, the expression

\begin{equation}\label{3}
(C_1A_2)(C_1A_2)^T-A_1(C_1\Sigma_{TI}\Sigma_{II}^{-1})^T-
(C_1\Sigma_{TI}\Sigma_{II}^{-1})A_1^T
\end{equation}
is monotonically decreasing. Since $A_1$ is bounded by Theorem 7, and $A_2A_2^T$ is nonnegative, the expression is convergent. In particular, $C_1A_2$ is also bounded along the flow. By the ~\eqref{UU}, both $A_1$ and $C_1A_2$ are $L^2$ integrable. Thus in fact we have pointwise convergence of $C_1A_2A_1$. 
Since $C_1$ may not be full rank, we call it partial convergence. If $C_1$ has full rank (which in general is the case of interest), then as $C_1A_2A_1$ is convergent, so is $A_2A_1$.

When does partial convergence imply the convergences of the solution $(A_1(t),A_2(t))$? The following result gives a sufficient condition.

\par\null
\noindent
{\bf Theorem 9:}
If the set of matrices $A_1,A_2$ 
satisfying:

\begin{align}\label{algebraic}
\begin{split}
& C_1A_2A_1=C_1\Sigma_{TI}\Sigma_{II}^{-1}\\
& (C_1A_2)+(C_1A_2)^T-A_1A_1^T=K\\
&C_1A_2A_2^TC_1^T+A_1(C_1\Sigma_{TI})^T+C_1\Sigma_{TI}A_1^T=L
\end{split}
\end{align}
is discrete, then $A_1(t$) and $C_1A_2(t)$ are convergent. 

\par \null
\noindent
{\bf Proof.} By the proof of Theorem 8, we know that $A_1(t)$, $C_1A_2(t)$ are bounded, and the limiting points of the pair $(A_1(t), C_1A_2(t))$ satisfy the relationships in Equation \ref{algebraic}. If the set is discrete, then the limit must be be unique and $A_1(t)$ and $C_1A_2(t)$ converge.\\

If $C_1$ has full rank, then the system in Equation \eqref{eq:LGN3} is convergent, if the assumptions in Theorem 9 are satisfied. Applying this result to the $\mathcal A[1,N,1]$ and $\mathcal A[N,1,N]$ cases, provides alternative proofs for Theorem 3 and Theorem 6. The details are omitted. Beyond these two cases, the algebraic set defined by Equation \eqref{algebraic} is quite complicated to study. The first non-trivial case 
that can be analyzed corresponds to the 
$\mathcal A[2,2,2]$ architecture. In this special case, we  we can solve the convergence problem entirely as follows. \\

For the sake of simplicity, we assume that $\Sigma_{II}=\Sigma_{TI}=C_1=I$. Then the system associated with Equation \eqref{eq:LGN3} can be simplified to:
 
\be
\begin{cases}
\frac{dB}{dt}= (I- BA)    A^t\\
\frac{dA}{dt}=  (I-BA)
\end{cases}
\label{eq:LGN3-2}
\ee
where $A(t), B(t)$ are $2\times 2$ matrix functions. By Theorem 7, we know that $B(t)A(t)$ is convergent. In order to prove that $B(t)$ and $A(t)$ are individually convergent, we prove the following result.

\par\null
\noindent
{\bf Theorem 10:}
Let $\mathcal F$ be the set of $2\times 2$ matrices $A,B$ satisfying the equations:
\begin{equation}\label{7-1}
\begin{split}
&B+B^T-AA^T=K\\
&A+A^T-BB^T=L\\
&AB=I
\end{split}
\end{equation}
where $K,L$ are fixed matrices. Then $\mathcal F$ is a discrete set and the system defined by Equation \ref{eq:LGN3-2} is convergent.

\par\null
\noindent {\bf Proof:} The proof is somewhat long and technical and thus is given in the Appendix. It uses basic tools from algebraic geometry.

\par\null\par

Theorem 10 provides evidence that in general the algebraic set defined by Equation \eqref{algebraic} might be discrete. Although at this moment we are not able to prove  discreteness in the general case, this is a question of separate interest in mathematics (real algebraic geometry).
The system defined by Equation \eqref{algebraic}
is an over-determined system of algebraic equations. For example, if $A(t), B(t)$ are $n\times n$ matrices, and if $C$ is non-singular, then the system contains $n(n+1)$ equations with $n^2$ unknowns. One can define the Koszul complex \cite{Eisenbud95} associated with these equations 
Using the complex, given specific matrices $C, \Sigma_{TI}, \Sigma_{II}, K, L$,  there is a constructive algorithmic way to determine whether the set is discrete. If it is, then the corresponding system of ODE is convergent.
\footnote{We thank Professor Vladimir Baranovsky for providing this information.}

\subsection{A Non-Linear Case}

As can be expected, the case of non-linear networks is 
challenging to analyze mathematically. In the linear case, the transfer functions are the identity and thus all the derivatives of the transfer functions are equal to 1 and thus play no role. The simulations reported above provide evidence that in the non-linear case the derivatives of the activation functions play a role in both RBP and SRBP.
Here we study a very simple non-linear case which provides some further evidence.

We consider a simple ${\cal A}[1,1,1]$ architecture, with a single power function non linearity with power $\mu \not = 1$ in the hidden layer, so that $O^1(S)=(S^1)\mu$. The final output neuron is linear $O^2(S^2)=S^2$ and thus the overall input-output relationship is:  $O=a_2(a_1I)^\mu$.
Setting $\mu$ to $1/3$, for instance, provides an S-shaped transfer function
for the hidden layer, and setting $\mu=1$ corresponds to the linear case analyzed in a previous section. The weights are $a_1$ and $a_2$ in the forward network, and $c_1$ in the learning channel. 

\par\null
\noindent
{\bf Derivation of the System Without Derivatives:}

When no derivatives are included, one obtains:
\be
\begin{cases}
\frac{da_2}{dt}=a_1^\mu [E(TI^\mu)-a_2a_1^\mu E(\mu I^2)]=a_1^\mu(\alpha -\beta a_2a_1^{\mu})
\\
\frac{da_1}{dt}=c_1[E(TI)-a_2a_1^\mu E(I^{\mu+1})]=c_1(\gamma -\delta a_2a_1^\mu )
\end{cases}
\label{eq:NLnof}
\ee
where here $\alpha=E(TI^\mu)$, $\beta=E(I^{2\mu})$, $\gamma=E(TI)$, and $\delta=E(I^{\mu +1})$. 
Except for trivial cases, such a system cannot have fixed points since in general one cannot have $a_2a_1^\mu=\alpha/\beta$ and 
$a_2a_1^\mu=\gamma/\delta$ at the same time.

\par\null
\noindent
{\bf Derivation of the System With Derivatives:}

In contrast, when the derivative of the forward activation is included the system becomes:

\be
\begin{cases}
\frac{da_2}{dt}= a_1^\mu [E(TI^\mu)-a_2a_1^\mu E(I^{2\mu})]=
a_1^\mu(\alpha -\beta a_2a_1^{\mu})
\\
\frac{da_1}{dt}=c_1\mu a_1^{\mu-1}E(TI^\mu)-a_2c_1 \mu    a_1^{2\mu-1}E(I^{2\mu})=a_1^{\mu-1}c_1 \mu
(\alpha-\beta a_2a_1^\mu) 
\end{cases}
\label{eq:wd1}
\ee
This leads to the coupling:

\be
a_1\frac{da_1}{dt}=c_1\mu \frac{da_2}{dt} \quad
{\rm or} \quad a_2=\frac{a_1^2}{2c_1\mu} + K
\label{eq:wd2}
\ee
excluding as usual the trivial cases where $c_1=0$ or $\mu=0$. Here $K$ is a constant depending only on $a_1(0)$ and $a_2(0)$. 
The coupling shows that if $da_1/dt=0$ then 
$da_2/dt=0$ and therefore in general limit cycles are not possible.
The critical points are given by the equation:

\be
\alpha -\beta a_2a_1^\mu=0 \quad {\rm or} \quad
a_2=\frac{\alpha}{\beta a_1^\mu}
\label{eq:critical100}
\ee
and do not depend on the weight in the learning channel.
Thus, in the non-trivial cases, $a_2$ is an hyperbolic function of $a_1^\mu$.
It is easy to see, at least in some cases, that the system converges to a fixed point. For instance, when $\alpha>0$, $c_1>0$, $\mu>1$, and $a_1(0)$ and $a_2(0)$ are small and positive, then $da_1/dt>0$ and $da_2/dt>0$ and both derivatives are monotonically increasing and 
$\alpha - \beta a_2a_1^\mu$ decreases monotonically until convergence to a critical point.{\it Thus in general the system including the derivatives of the forward activations is simpler and better behaved}. In fact, we have a more general theorem.

\par\null
\noindent 
{\bf Theorem 12:} 
Assume that $\alpha>0$, $\beta>0$ $c_1>0$, and $\mu \geq 1$. Then for any positive initial values $a_1(0) \geq 0$ and 
$a_2(0) \geq 0$, the system described by Equation \ref{eq:wd1}
is convergent to one of the positive roots of the equation for $t$:

\be
\alpha -\beta(\frac{t^2}{2c_1\mu} + K)t^\mu=0
\label{eq:110}
\ee

\par\null
\noindent 
{\bf Proof:}
Using Equation \ref{eq:wd2}, the differential equation for $a_1$ can be rewritten as:

\be
\frac{da_1}{dt}=\mu a_1^{\mu-1}c_1(\alpha - \beta(\frac{a_1^2}{2c_1 \mu}+K)a_1^\mu)=Q(a_1)
\label{eq:111}
\ee
When $\mu$ is an integer, $Q(a_1)$ is a polynomial of odd degree with a leading coefficient that is negative and therefore, using Theorem 1, the system is convergent.
If $\mu$ is not an integer, let $r_1 < \ldots <r_k$ be the positive roots of the function $Q$. The proof then proceeds similarly to the proof of Theorem 1.
That is this differential equation (Equation 
\ref{eq:111}) is convergent to one of the (non-negative) roots of $Q(t)$. However, since $a_1(0)>0$, a more careful analysis shows that it is not for $a_1$ to converge to zero. Thus $a_1$ must converge to a 
positive root of Equation \ref{eq:110}.

\par\null
\noindent
{\bf Gradient Descent Equations:}
Finally, for comparison, in the case of gradient descent, the system is given by:

\be
\begin{cases}
\frac{da_2}{dt}= a_1^\mu [E(TI^\mu)-a_2a_1^\mu E(I^{2\mu})]=
a_1^\mu(\alpha -\beta a_2a_1^{\mu})
\\
\frac{da_1}{dt}=a_2\mu a_1^{\mu-1}E(TI^\mu)-a_2^2 \mu    a_1^{2\mu-1}E(I^{2\mu})=a_1^{\mu-1}a_2 \mu
(\alpha-\beta a_2a_1^\mu) 
\end{cases}
\label{eq:}
\ee
Except for trivial cases, the critical points are again given by Equation \ref{eq:critical100}, and the system always converges to a critical point.

\section{Conclusion}

Training deep architectures with backpropagation on digital computers is useful for practical applications, and it has become easier than ever, in part because of the creation of software packages with automatic differentiation capabilities. This convenience, however, can be misleading as it hampers thinking about the constraints of learning in physical neural systems, which are merely being mimicked on digital computers.
Thinking about learning in physical systems is useful in many ways: it leads to the notion of local learning rules, which in turn identifies two fundamental problems facing backpropagation in physical systems. First backpropagation is not local, and thus a learning channel is required to communicate error information from the output layer to the deep weights. Second, backpropagation requires symmetric weights, a significant challenge for those physical systems that cannot use the forward channel in the reverse direction, thus requiring a different pathway to communicate errors to the deep weights.

RBP is one mode for communicating information over the learning channel, that completely bypasses the need for symmetric weights, by using fixed random weights instead. However RBP is only one possibility among many other ones for harnessing randomness in the learning channel. Here we have derived several variants of RBP and studied them through simulations and mathematical analyses.
Additional variants are studied in a followup paper
\cite{baldiSymmetries2017} which considers additional symmetry issues such as having a learning channel with an architecture that is not a symmetric version of the forward architecture, or having non-linear units in the learning channel that are similar to the non-linear units of the forward architecture.

In combination, the main emerging picture is that the general concept of RBP is remarkably robust as most of the variants lead to robust learning. RBP and its many variants 
do not seem to have a practical role in digital simulations as they often lead to slower learning, but they should be useful in the future both to better understand biological neural systems, and to implement new neural physical systems in silicon or other substrates.

\begin{table}[h!]
\renewcommand{\arraystretch}{1.5}
\begin{center}
    \begin{tabular}{ | c | l |   }
    \hline
 \textbf{Information} &  \textbf{Algorithm}   \\ \hline
 $I^{h}_{ij}=I^{h}_{ij}(T, O,w_{rs}^l (l>h),f' (l \geq h))$ & General Form\\ \hline
 $I^{h}_{ij}=I^{h}_{i}(T, O, w_{rs}^l (l>h),f' (l \geq h))$ & BP (symmetric weights)  \\ \hline
$I^{h}_{ij}=I^{h}_{i}(T-O, w_{rs}^l (l>h),f' (l \geq h))$  & BP (symmetric weights) \\\hline
$I^{h}_{ij}=I^{h}_{i}(T-O, w_{rs}^l (l>h+1),w^{h+1}_{ki},f' (l \geq h))$  & BP (symmetric weights) \\\hline
 $I^{h}_{ij}=I^{h}_{i}( T-O,  r_{rs}^l (l\geq h+1),r^{h}_{ki},f' (l \geq h))$  & RBP (random weights) \\ \hline
 $I^{h}_{ij}=I^{h}_{i}( T-O, r^h_{ki},f' (l \geq h))$  &  SRBP (random skipped weights) \\ \hline
  $I^{h}_{ij}=I^{h}_{i}( T-O, r^h_{ki},f' (l =h))$  & SRBP  (random skipped weights)\\ \hline
  $I^{h}_{ij}=I^{h}_{i}( F(T-O), f' (l =h))$  &  F sparse/low-prec./adaptive/non-lin.\\ \hline  
    \end{tabular}
\end{center}
\caption{Postsynaptic information required by deep synapses for optimal learning. $I^h_{ij}$ represents the  signal carried by the deep learning channel and the postsynaptic term in the learning rules considered here. Different algorithms reveal the essential ingredients of this signal and how it can be simplified. In the last row, the function $F$ can be implemented with sparse or adaptive matrices, carry low precision signals, or include non-linear transformations in the learning channel (see also 
\cite{baldiSymmetries2017}).}
    \label{tab:channel}
\end{table}

In supervised learning, the critical equations show that in principle any deep weights must depend on all the training examples and all the other weights of the network. Backpropagation shows that it is possible to derive effective learning rules of the form
$\Delta w_{ij}^h=\eta I_{ij}^h O_j^{h-1}$ where the role of the lower part of the network is subsumed by the presynaptic activity term $O_j^{h-1}$ and $I_{ij}^h$ is a signal communicated through the deep learning channel that carries information about the outputs and the targets to the deep synapses. Here we have studied what kind of information must be carried by the signal $I_{ij}^h$ and how much it can be simplified 
(Table \ref{tab:channel}). The main conclusion is that the postynaptic terms must: (1) implement gradient descent for the top layer (i.e. random weights in the learning channel for the top layer do not work at all); (2) for any other deep layer $h$ it should be of the form $f'F(T-O)$, where$f'$ represents the derivatives of the activations of the units in layer $h$ (the derivatives above are not necessary) and $F$ is some reasonable function of the error $T-O$. By reasonable, we mean that the function $F$ can be linear, or a composition of linear propagation with non-linear activation functions, it can be fixed or slowly varying, and when matrices are involved these can be random, sparse, etc.
As can be expected, it is better if these matrices are full rank although gracious degradation, as opposed to catastrophic failure, is observed when these matrices deviate slightly from the full rank case. 

The robustness and other properties of these algorithms cry for explanations and more general principles. We have provided both intuitive and formal explanations for several of these properties. On the mathematical side, polynomial learning rules in linear networks lead to systems of polynomial differential equations. We have shown in several cases that the corresponding ODEs converge to an optimal solution. However these polynomial systems of ODEs rapidly become complex and, while the results provided are useful, they are not yet complete, thus providing directions for future research.

\section*{Acknowledgement}

Work supported in part by NSF grant 
IIS-1550705 and a Google Faculty Research Award to PB,
and NSF grant DMS-1547878 to ZL. 
We are also grateful for a hardware donation from NVDIA Corporation. 

\section*{Appendix: Proof of Theorem 10}

Assume that $(A,B)\in\mathcal F$. If near $(A,B)$, $\mathcal F$ is not discrete, then there are real analytic matrix-valued functions $(A(t), B(t))\in\mathcal F$ for small $t>0$ such that $(A(0),B(0))=(A,B)$.  Moreover, if we write:
\be
A(t)=A+tE+\frac{t^2}{2}F+\frac{t^3}{6}G+o(t^3)
\ee
then $E\neq 0$.
We  use $A',A'', A''', B',B'',B'''$ to denote $A'(0), A''(0), A'''(0), B'(0),$
$ B''(0), B'''(0)$, respectively. 
The general strategy is to prove that $E=0$ or, in the case $E\neq 0$, to take higher order derivatives to reach a contradiction. 

It is easy to compute:

\be
B'=-A^{-1}A'A^{-1}=-BEB
\ee
By taking the derivative of the first two relations in Equation \eqref{7-1}, we have:

\be
\begin{split}
&-BEB-(BEB)^T-EA^T-AE^T=0\\
&E+E^T+BEBB^T+B(BEB)^T=0
\end{split}
\ee
Let:

\be
X=EA^T+BEB \quad 
Y=E+BEBB^T
\ee
Then by the above equations, both $X,Y$ are  skew symmetric, and 
we have $YA^T=X$.
If $Y\neq 0$, using an orthogonal transformation and  scaling, we may assume that:
 
\be
Y=\begin{bmatrix}0&-1\\1&0\end{bmatrix}
\ee
Write: 
\begin{equation}\label{A}
A=\begin{bmatrix}a&b\\c&d\end{bmatrix}
\end{equation}
Then:

\be
YA^T=\begin{bmatrix}-b&-d\\a&c\end{bmatrix}
\ee
Since $X$ skew-symmetric also, we must have $b=c=0$, and $a=d$. Thus
$A=a I$ for a real number $a\neq 0$. As a result, we have:

\be
K=(\frac 2a-a^2)I \quad L=(2a-\frac{1}{a^2})I
\ee
and $(A,B)=(a I,a^{-1}I)$.  Let $(\tilde A(t),\tilde B(t))$ be the upper triangular matrices obtained by orthogonal transformation from $(A(t), B(t))$. Since both $K,L$ are proportional to the identity,
$(\tilde A(t),\tilde B(t))\in\mathcal F$.
Now let us write:

\be
\tilde A(t)=\begin{bmatrix}\tilde a&\tilde b\\0&\tilde d\end{bmatrix}
\ee
Then 
the equation $B+B^T-AA^T=K$  is equivalent to the following system:

\be
\left\{
\begin{array}{l}
2\tilde a^{-1}-(\tilde a^2+\tilde b^2)=2a^{-1}-a^2\\
2\tilde d^{-1}-\tilde d^2=2a^{-1}-a^2\\
-\tilde b\tilde a^{-1}\tilde d^{-1}-\tilde b\tilde d=0
\end{array}
\right\}
\label{eq:algsys10}
\ee
Since $t$ is  small, $\tilde A(t)$ should be sufficiently close to $aI$. From the second equation of the system above, we have $\tilde d=a$. If $\tilde b= 0$, then we conclude from the first equation of the same system that $\tilde a=a$, and hence $\tilde A(t)=aI$. This implies that $(A(t), B(t))=(A,B)$. So in this case $E=0$.

Things are more complicated when $\tilde b\neq 0$. We first assume that $a\neq -1$. In this case, from the third equation of the system above, we have $\tilde a^{-1}\tilde d^{-1}+\tilde d=0$. Since we already have $\tilde d=a\neq 1$, for sufficiently small $t$, $\tilde a=-\tilde d^{-2}=-a^{-2}$, which is distinct from $a$. Thus in this case $\tilde b$ must be zero. 
If $a=-1$, then we have $\tilde d=-1$ and $\tilde a=-1$. Using the first equation of the system above, we have $\tilde b=0$ and the again $(A(t), B(t))=(\tilde A(t), \tilde B(t))=(A,B)$, and we conclude that $E=0$. \\

From the results above, we know that if $Y\neq 0$ or if $A$ is proportional to the identity, near $(A, B)\in\mathcal F$, there are no other elements in $\mathcal F$ and thus $\mathcal F$ is discrete.  When $X=Y=0$, it is possible to have $E\neq 0$. However, we have the following Lemma:

\par\null
\noindent
{\bf Lemma 3:}
If $X=Y=0$, and if $A\neq -I$, then $E$ is not an invertible matrix. 

\par \null
\noindent
{\bf Proof.}
By contradiction, assume that $E$ is invertible. Then from $X=0$, we have:

\be
-A=EBB^TE^{-1}
\ee
By taking determinant on both sides, we get:

\be
\det A=\det (-A)=\det (BB^T)
\ee
Thus we have:

\be
\det A=1
\ee
Since $A$ is similar to a negative definite matrix $-BB^T$, the eigenvalues $\lambda_1,\lambda_2$ of $A$ are all negative. Since $\lambda_1\lambda_2=\det A=1$, we have:

\be
-\lambda_1-\lambda_2\geq 2
\ee
Using the same matrix representation as in Equation \eqref{A}, we have:

\be
-a-d=-{\rm Tr}\,A={\rm Tr}\,(BB^T)=a^2+b^2+c^2+d^2
\ee
However:

\be
a^2+b^2+c^2+d^2=(a+d)^2+(b-c)^2-2\geq (a+d)^2-2\geq -a-d,
\ee
and the equality is true if and only if $b=c$ and $a+d=-2$. Since $-\lambda_1-\lambda_2=2$ and $\lambda_1\lambda_2=1$,  the eigenvalues of $A$ must be $-1,-1$, which implies $b=c=0$.  Thus $A=-I$ which is impossible by our assumption.  

Next we consider the remaining case: $X=Y=0$, and $E$ is not invertible (but not equal to zero), and $A$ is not proportional to the identity.  In this case, we have to take up to third order derivatives to reach the conclusion.
By taking derivatives of the first two relations in Equation \eqref{7-1}, we get:

\be
P+P^T=Q+Q^T=R+R^T=S+S^T=0
\ee
where:

\begin{align}
\begin{split}
&P=-B''+A''A^T+A'(A')^T\\
&Q=A''-B''B^T-B'(B')^T\\
&R=-B'''+A'''A^T+3A''(A')^T\\
&S=A'''-B'''B^T-3B''(B')^T
\end{split}
\end{align}
Similar to the relations between the matrices $X,Y$, we have:
\be
QA^T-P=-B'(B')^TA^T-A'(A)^T
\ee
Since $AB=I$, we have:
 
\begin{equation}\label{7}
(B')^TA^T=-B^T(A')^T
\end{equation}
Thus:

\be
QA^T-P=B'B^TA^T-EA^T=-BEBB^TA^T-EA^T=0
\ee
because $X=0$.
Since $A$ is not proportional to the identity, then we must have $P=Q=0$ as in the case for $X$ and $Y$.

The relationship between $R,S$ is more complicated, but can be computed using the same idea. We first have:

\be
SA^T-R=-3B''(B')^TA^T-3A''(A')^T
\ee
Using Equation \eqref{7} and the fact that $P=0$, we have:

\be
SA^T-R=-3A'(A')^TB^T(A')^T=3EE^TB^TE^T
\ee
Since $E$ is not invertible and we assume that $E\neq 0$, we must have:

\be
E=\xi\eta^T
\ee
for some column vectors $\xi, \eta$. From the fact that $Y=0$, we conclude that:
 
\be
\xi\eta^T+B\xi\eta^TBB^T=0
\ee
and:

\be
B\xi=-\frac{\|\eta\|^2}{\|B^T\eta\|^2}\xi
\ee
Thus we compute:
\be
EE^TB^TE^T=-\frac{\|\eta\|^4\cdot \langle \xi,\eta\rangle}{\|B^T\eta\|^2}\xi\xi^T
\ee
and:

\be
SA^T-R=-3\frac{\|\eta\|^4\cdot \langle \xi,\eta\rangle}{\|B^T\eta\|^2}\xi\xi^T
\ee
If $\langle \xi,\eta\rangle\neq 0$, then $S\neq 0$. Thus:

\be
A^T=S^{-1}R-3\frac{\|\eta\|^4\cdot \langle \xi,\eta\rangle}{\|B^T\eta\|^2}S^{-1}\xi\xi^T
\ee
For the matrix $S^{-1}\xi\xi^T$, both the trace and determinant are zero. So the eigenvalues  are zero. On the other hand, since both $S,R$ are skew-symmetric matrices, $S^{-1}R$ is proportional to the identity. As a result, the matrix $A^T$, hence $A$, has two identical eigenvalues. Let $\lambda$ be an eigenvalue of $A$, then:

\be
{\rm Tr}\, (A)=2\lambda,\quad \det(A)=\lambda^2
\ee
Taking the trace in the first two relations of Equation\eqref{7-1}, we get:

\be
\begin{split}
& 4\lambda^{-1}-\|A\|^2={\rm Tr}\,(K)\\
& 4\lambda-\lambda^{-2}\|A\|^2={\rm Tr}\,(L)
\end{split}
\ee
Thus for fixed $K,L$, $\lambda$ and $\|A\|$ can only assume discrete values. Since $t$ is small,  $A(t)=Q(t)AQ(t)^T$ for some orthogonal matrix $Q(t)$. Let us write:

\be
A=\begin{bmatrix} \lambda& b\\0&\lambda\end{bmatrix},\quad Q'(0)=\begin{bmatrix} 0&-1\\-1&0\end{bmatrix}
\ee
Then $E=A'(0)$ is equal to:
\be
E=\begin{bmatrix} b&0\\0&-b\end{bmatrix}
\ee
By Lemma 3, $E$ is not invertible. Thus $b=0$. But if $b=0$, then $A$ is proportional to the identity and this case has been discussed above.  

We must still deal with the case $\langle\xi,\eta\rangle=0$. Without loss of generality, we may assume that:

\be
\xi=\begin{bmatrix}1\\0\end{bmatrix},\quad \eta=
\begin{bmatrix}0\\1\end{bmatrix}
\ee
By checking the equation $AE=-EBB^T$, we can conclude that:
 
\be
A=\begin{bmatrix}-d^{-2}&0\\0&d\end{bmatrix}.
\ee
In fact, when $t$ is small, the eigenvalues of $A(t)$ must be $-d^{-2}$ and $d$ for some $d\neq 0$. Again, by taking the trace of the first two relations in Equation \eqref{7-1}, we get:

\be
\begin{split}
&-2d^2+2d^{-1}-\|A\|^2={\rm Tr}\,(K);\\
&-2d^{-2}+2d-d^2\|A\|^2={\rm Tr}\,(L).
\end{split}
\ee
Therefore, $d$ is locally uniquely determined by $K,L$. Finally, if we write $A(t)=Q(t)AQ(t)^T$ and assume that:
 
\be
Q'(0)=\begin{bmatrix} 0&-1\\1&0\end{bmatrix}, 
\ee
we have:

\be
E=\begin{bmatrix} 0&d+d^{-2}\\d+d^{-2}&0\end{bmatrix}.
\ee
Since $E$ must be singular, we have $d=-1$ and hence $A=-I$. This case has been covered above and thus the proof of Theorem 10 is complete.

\bibliographystyle{abbrv}
\bibliography{baldi,nn,AWnetsbib}

\end{document}